\documentclass[conference]{IEEEtran}
\usepackage{times}

\usepackage[numbers]{natbib}
\usepackage{multicol}
\usepackage[bookmarks=true]{hyperref}
\usepackage{authblk}
\usepackage{graphics} 
\usepackage{epsfig} 
\usepackage{mathptmx} 
\usepackage{times} 
\usepackage{amsmath} 
\usepackage{amssymb}  
\usepackage{booktabs}
\usepackage{float}
\usepackage{multirow}
\usepackage{caption}
\usepackage{newtxtext, newtxmath}
\usepackage{tabularx}
\usepackage{longtable}
\usepackage{hyperref}

\usepackage{flushend}
\usepackage{stfloats}
\usepackage{balance}
\usepackage{xspace}
\usepackage{listings}
\usepackage{algorithm}
\usepackage{algpseudocode}
\usepackage{tcolorbox}
\usepackage{graphicx}

\newcolumntype{M}[1]{>{\centering\arraybackslash}m{#1}}

\title{\LARGE \bf
MotionTrans: Human VR Data Enable Motion-Level Learning for Robotic Manipulation Policies
}

\author{
\textbf{Chengbo Yuan\textsuperscript{1,2}, 
Rui Zhou*\textsuperscript{5}, 
Mengzhen Liu*\textsuperscript{3}, 
Yingdong Hu\textsuperscript{1,2}, 
Shengjie Wang\textsuperscript{1,2}} \\
\textbf{
Li Yi\textsuperscript{1,2},  
Chuan Wen\textsuperscript{4}, 
Shanghang Zhang\textsuperscript{3}, 
Yang Gao\textsuperscript{1,2}$^\dag$}\\
\textsuperscript{1}Institute for Interdisciplinary Information Sciences, Tsinghua University\ \ 
\textsuperscript{2}Shanghai Qi Zhi Institute \\ 
\textsuperscript{3}State Key Laboratory of Multimedia Information Processing, School of Computer Science, Peking University\\
\textsuperscript{4}Shanghai Jiao Tong University\ \ 
\textsuperscript{5}Wuhan University\\
\vspace{1mm}
*\ Indicates equal contribution. \dag\ The corresponding author.\\
\href{https://motiontrans.github.io/}{\large\textbf{https://motiontrans.github.io/}}
}

\begin{document}

\newcommand{\insertteaser}{
    \includegraphics[width=\linewidth]{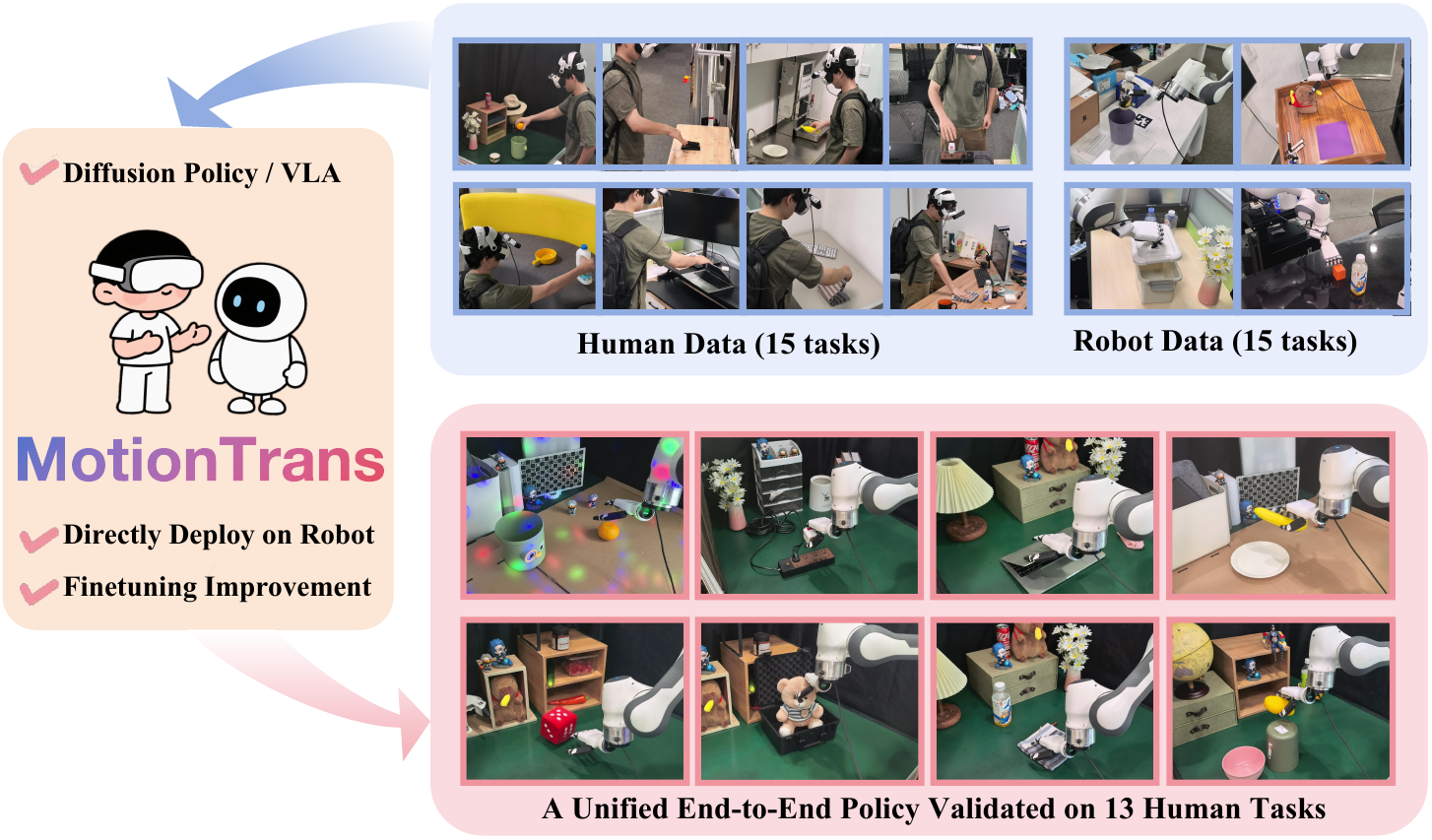}
    \captionof{figure}{We propose \textit{MotionTrans}, a framework that enables \textbf{motion-level} learning from VR-collected human data. By cotraining on 15 human tasks and 15 robot tasks, we empower end-to-end robotic manipulation policies to directly perform tasks in human data on real robot hardware. Our framework also improves finetuning performance when a few robot demonstrations are available for these tasks.}
    \vspace{-2mm}
    \label{fig:method-overview}
}

\makeatletter
\apptocmd{\@maketitle}{\centering\insertteaser}{}{}
\makeatother
\maketitle
\setcounter{figure}{1}

\thispagestyle{empty}
\pagestyle{empty}

\begin{abstract}

Scaling real robot data is a key bottleneck in imitation learning, leading to the use of auxiliary data for policy training. 
While other aspects of robotic manipulation such as image or language understanding may be learned from internet-based datasets, acquiring motion knowledge remains challenging. 
Human data, with its rich diversity of manipulation behaviors, offers a valuable resource for this purpose.
While previous works show that using human data can bring benefits, such as improving robustness and training efficiency, it remains unclear whether it can realize its greatest advantage: \textit{enabling robot policies to directly learn new motions for task completion}.
In this paper, we systematically explore this potential through multi-task human-robot cotraining. We introduce \textit{MotionTrans}, a framework that includes a data collection system, a human data transformation pipeline, and a weighted cotraining strategy. By cotraining 30 human-robot tasks simultaneously, we direcly transfer motions of 13 tasks from human data to deployable end-to-end robot policies.
Notably, 9 tasks achieve non-trivial success rates in zero-shot manner.
\textit{MotionTrans} also significantly enhances pretraining-finetuning performance (+40\% success rate).
Through ablation study, we also identify key factors for successful motion learning: cotraining with robot data and broad task-related motion coverage. 
These findings unlock the potential of motion-level learning from human data, offering insights into its effective use for training robotic manipulation policies.
All data, code, and model weights are open-sourced \href{https://motiontrans.github.io/}{https://motiontrans.github.io/}.

\end{abstract}

\section{Introduction}

Learning robotic manipulation policies from teleoperated demonstrations has progressed rapidly in recent years~\cite{cheng2024opentelevision, chi2023diffusion, black2024pi_0}. However, collecting large-scale robot datasets remains costly and labor-intensive~\cite{khazatsky2024droid, o2024openxembodiment}, creating a significant bottleneck for further improvement of manipulation abilities. To address data scarcity, researchers have turned to auxiliary sources, such as images or language~\cite{ji2025robobrain, zhou2025roborefer} to help policy training. While internet data provides abundant vision-language knowledge to aid policy learning~\cite{intelligence2025pi_05}, acquiring motion knowledge remains a significant challenge.

Human data~\cite{qiu2025humanoidhuman, hoque2025egodex} represents a particularly promising source to solve this: it is abundant, easy to collect, and rich in diverse manipulation behaviors~\cite{hoque2025egodex}. Previous works have leveraged human demonstrations to extract task-aware representations, such as affordances~\cite{bahl2023affordances} or keypoint flows~\cite{yuan2024generalflow}, to support motion transfer. However, the introduction of intermediate representation hinders integration with mainstream end-to-end policies. More recently, with advances in wearable sensing, researchers begin to explore the use of human motion data (with hand poses recorded from VR device) directly for robot policy cotraining or pretraining~\cite{kareer2024egomimic, qiu2025humanoidhuman, yang2025egovla, luo2025being, bi2025hrdt}. These approaches have shown benefits for visual grounding~\cite{luo2025being}, robustness~\cite{yang2025egovla} and training efficiency~\cite{bi2025hrdt}. However, it is still uncertain whether it can fully realize its greatest advantage: \textit{allowing robot policies to directly acquire new motions for task completion}.

In this paper, we investigate this question by introducing \textit{MotionTrans}, a framework designed to \textbf{directly learn 13 robot-executable motions from human data for a unified, end-to-end robot policy.} This is achieved through multi-task human-robot cotraining. We develop a VR-based teleoperation system and data collection pipeline to construct the \textit{MotionTrans Dataset}, which includes 3,213 demonstrations across 15 human tasks and 15 robot tasks from more than 10 scenes. We further propose a transformation procedure that maps human demonstrations into the robot’s observation–action space, making them compatible with mainstream end-to-end policies such as Diffusion Policy~\cite{chi2023diffusion} or the Vision-Language-Action model ($\pi_0$-VLA)~\cite{black2024pi_0}. Finally, we adopt a weighted cotraining strategy that jointly optimizes over both human and robot tasks. We name the entire framework \textit{MotionTrans} because it enables motion transfer from human data to deployable robot policies.


We first evaluate the zero-shot performance on all human tasks. This means that we directly deploy policies to robot without collecting any robot data for these tasks. Results show that Diffusion Policy~\cite{chi2023diffusion} and $\pi_0$-VLA model~\cite{black2024pi_0} achieve non-trivial success rates for 9 tasks in total. Even in unsuccessful cases, they exhibit meaningful motion for task completion, such as reaching target objects. 
We also find that, when few robot demonstrations of these human tasks are available for finetuning, pretraining on the \textit{MotionTrans Dataset} leads to an average 40\% boost in success rate on these tasks. Further analysis indicates that the effectiveness of motion transfer depends on the presence of both robot demonstrations and sufficient task-related motion coverage during training. Together, these findings highlight the possibility for motion-level learning from human data, and provide a clear framework and principles for achieving this.
Our contributions can be summarized as:
\begin{itemize}
    \item \textit{MotionTrans}, a framework for end-to-end human-to-robot motion transfer, including data collection system, a pipeline to transform human data into robot format, and a weighted human-robot cotraining strategy.
    \item \textit{MotionTrans Dataset}, containing 3,213 demonstrations for 15 human tasks and 15 robot tasks across 10+ scenes.
    \item \textbf{\textit{MotionTrans} enables explicit human motions transfer for end-to-end robot policies, even for zero-shot settings} (directly learn 13 tasks from human data).
    \item Key factors for successful motion transfer: robot data cotraining and sufficient task-related motion coverage. 
\end{itemize}

\section{Related Work}

\subsection{Imitation Learning for Robot Manipulation}

Imitation learning~\cite{lin2024datascaling,barreiros2025careful,shi2025diversity, liu2024robomamba} has made significant progress in recent years. By learning motion from training data~\cite{cheng2024opentelevision, chen2024arcap}, imitation policies can effectively perform a wide range of manipulation tasks~\cite{chi2023diffusion, zhao2024aloha}, including challenging multi-task settings~\cite{zitkovich2023rt2, liu2024rdt, black2024pi_0, bi2025hrdt, liu2025hybridvla}. In this paper, we focus on two widely-used architectures for imitation learning: Diffusion Policy~\cite{chi2023diffusion} and the $\pi_0$ Vision-Language-Action Model ($\pi_0$-VLA)~\cite{black2024pi_0}. However, the scalability of training data remains a major challenge, due to the high cost of collecting real-robot data~\cite{o2024openxembodiment, khazatsky2024droid, wu2024robomind}.
This has led to the use of auxiliary data~\cite{ji2025robobrain, zhou2025roborefer,liu2024segment} for policy training. Despite ability such as image or language understanding in robotic manipulation could improve from internet-based pretraining~\cite{intelligence2025pi_05, lin2025onetwovla}, acquiring motion knowledge remains difficult. Human data~\cite{hoque2025egodex, liu2022hoi4d, yuan2024egomono4d, damen2022epic}, with its abundant and diverse manipulation behaviors, provides a valuable supplement for this.

\subsection{Task-Aware Representation Learning from Human}

Early works have leveraged task-aware representations for human-to-robot knowledge transfer. Self-supervised learning has been used for implicit task-aware representations~\cite{nair2022r3m, karamcheti2023voltron, majumdar2023vc1, ye2024lapa, bu2025univla} learning, while representations like affordances~\cite{bahl2023affordances, kuang2024ram, shi2025zeromimic}, object poses~\cite{hsu2024spot}, videos~\cite{bharadhwaj2024gen2act, patel2025roboticvido}, and motion flows~\cite{yuan2024generalflow, wen2023any, xu2024flow2act, ren2025motiontrack} support motion-aware representation learning. Some approaches use wrist trajectories as prompts for one-shot human-to-robot skill transfer~\cite{kim2025uniskill, zhou2025yoto, zhu2025learninggenhumanprompt, tang2025functo, park2025demodiffusion}. EgoZero~\cite{liu2025egozero} predicts wrist poses from smart glasses, but relies on keypoint-based representations~\cite{wang2025skil} for policy observations. The use of intermediate representations in these methods limits their integration with mainstream end-to-end visuomotor policy learning~\cite{chi2023diffusion, black2024pi_0}, restricting their future applicability.

\subsection{End-to-End Policy Learning with Posed Human Data}

Human motion data can be captured through hand-held SLAM-based device~\cite{chi2024umi, xu2025dexumi}, but often limited to only wrist camera sensing~\cite{tao2025dexwild}. Recent advancements in wearable sensing~\cite{engel2023projectaria, chen2024arcap, qiu2025humanoidhuman} now allow easy collection of posed human data (with hand keypoints, wrist poses information etc.) through VR devices~\cite{hoque2025egodex}. This data provide action label for prediction, supporting end-to-end policy learning~\cite{lepert2025phantom}. Some studies cotrain human and robot data~\cite{kareer2024egomimic, qiu2025humanoidhuman, niu2025human2locoman, lepert2025masquerade, tao2025dexwild, liuimmimic}, while others first pretrain with human data and then finetune with robot demonstrations~\cite{yang2025egovla, luo2025being, bi2025hrdt}. These works have shown policy improvements in visual grounding~\cite{luo2025being}, robustness~\cite{qiu2025humanoidhuman, yang2025egovla}, and training efficiency~\cite{bi2025hrdt, kareer2024egomimic}. However, whether it can achieve direct transfer of motions from human to robot remains unclear~\cite{liu2025egozero}. To the best of our knowledge, our paper is the first to systematically verify motion-level end-to-end learning from human data.

\begin{figure*}[t]
    \centering
    \includegraphics[width=1.0\linewidth]{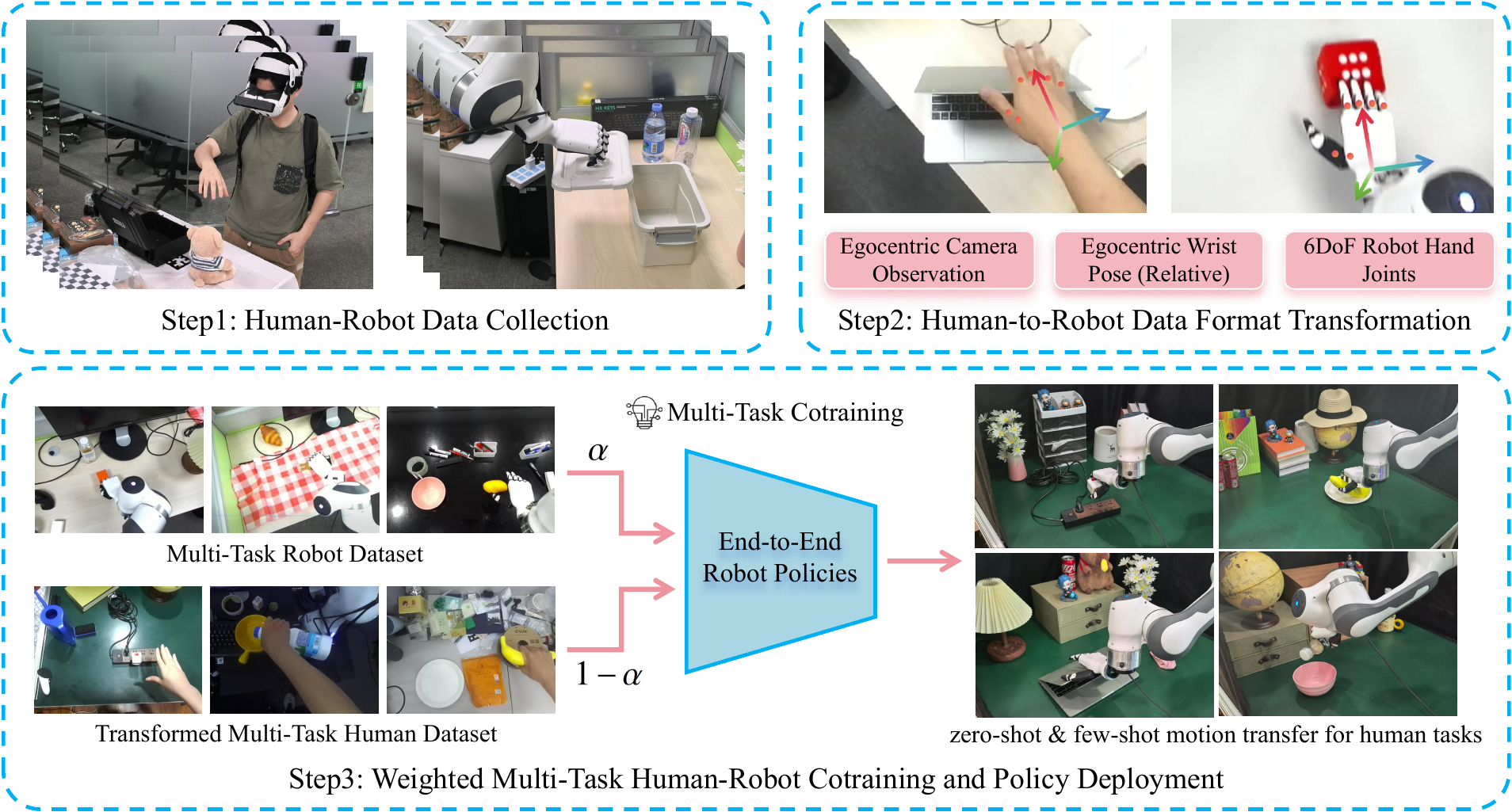}
    \caption{Illustration of our proposed \textit{MotionTrans} framework, which consists of a human-robot data collection system, a pipeline for transforming human data into robot format, and a weighted human-robot multi-task cotraining strategy. After training, we enable the direct deployment of the trained policies to perform tasks in human datasets on real robots.}
    \label{fig:main_method}
\end{figure*}

\section{MotionTrans}

In this section, we present our proposed \textit{MotionTrans} framework (Figure~\ref{fig:main_method}). The core idea is to first transform human data to robot data format, and then jointly learn from human and robot data within the robot observation-action space. By training policies in robot space, we can directly deploy policies to perform tasks from human data on real-world robots, i.e., enabling explicit human-to-robot motion transfer.
We first introduce the motion transfer problem and define the observation-action space of the policy (Section~\ref{sec:motiontrans_problem_define}). To facilitate human-robot data cotraining, we develop data collection systems for both human and robot data (Section~\ref{sec:motiontrans_data_collection}). We then propose a pipeline to convert human data into robot format (Section~\ref{sec:motiontrans_data_processing}). This ensures compatibility with mainstream robot policies, enabling subsequent end-to-end cotraining. Finally, we choose the architecture of robot policies and apply human-robot multi-task cotraining (Section~\ref{sec:motiontrans_cotraining}).

\subsection{Problem Definition}~\label{sec:motiontrans_problem_define}
Our goal is to enable explicit human-to-robot motion transfer. Considering the embodiment gap between human and robot~\cite{kareer2024egomimic}, we explore this problem within a multi-task human–robot cotraining framework, where robot data for certain tasks are available to help motions in human data adapt to the robot. Specifically, we aim to train a policy \( P_{\text{policy}} \) on \( D = D_{\text{robot}} \cup D_{\text{human}} \), where \( D_{\text{robot}} = \{ D^i_{\text{robot}} \mid i = 1, \ldots, N_{\text{robot}} \} \) is the robot dataset, and \( D_{\text{human}} = \{ D^i_{\text{human}} \mid i = 1, \ldots, N_{\text{human}} \} \) is the human dataset. Each \( D^i \) represents a sub-dataset corresponding to a specific task, and the task sets of the human and robot data are \textbf{non-overlapping}. 
After training, we deploy \( P_{\text{policy}} \) on a real-world robot and evaluate its performance on \textbf{tasks from \( D_{\text{human}} \)} to assess the effectiveness of motion transfer. This is defined as the \textbf{zero-shot} setting, since the evaluation tasks contain no corresponding robot data for training. We also evaluate the performance of \textbf{few-shot finetuning} setting, where a small number of robot demonstrations for the tasks from $D_{\text{human}}$ are available to further finetune \( P_{\text{policy}} \). 

\begin{figure}[t]
    \centering
    \includegraphics[width=1.0\linewidth]{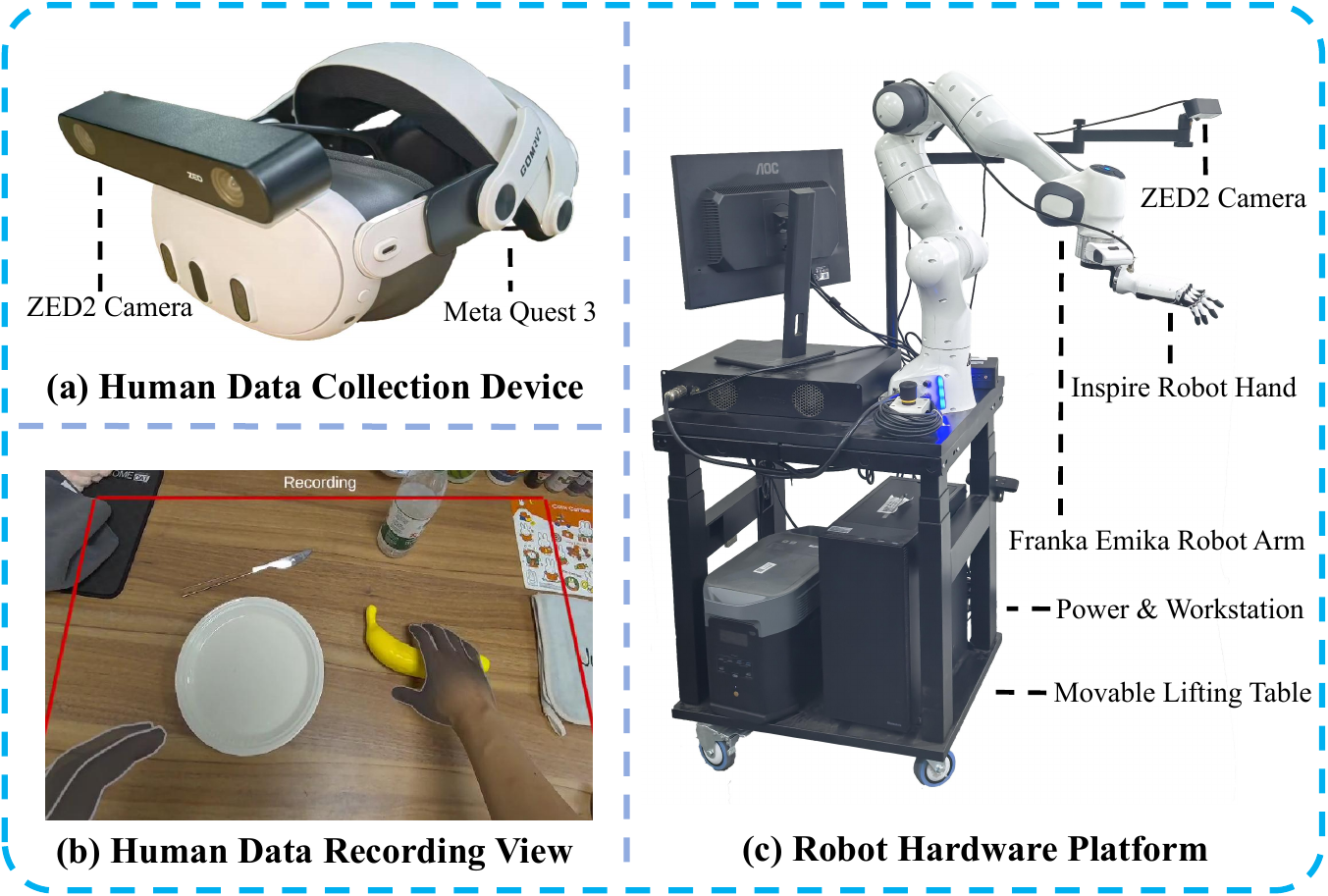}
    \caption{Illustration of our hardware system, which includes a human VR-based data collection device and a single-arm robot platform. A screenshot of the VR device during human data collection is also provided.}
    \label{fig:main_hardware}
\end{figure}

We define the input and output of our policies within the robot observation-action space \( S = (I_t, P_t, A_t) \). At each timestamp \( t \), the policy receives an egocentric RGB image \( I_t \in \mathbb{R}^{H \times W \times 3} \) and proprioceptive states \( P_t \in \mathbb{R}^{T_P \times D} \), where \( T_P \) is the history length and \( D \) is the state dimension. For simplicity, this work focuses on single-arm tasks (Figure~\ref{fig:main_dataset}), thus \( D \) corresponds to the concatenation of one robot wrist pose and one robot hand joint state (Figure~\ref{fig:main_hardware}(c)). The policy outputs an action chunk prediction \( A_t \in \mathbb{R}^{T_A \times D} \)~\cite{chi2023diffusion}, where \( T_A \) denotes the action prediction horizon. Next, we describe the details of our human-robot data collection and processing system.

\subsection{Human-Robot Data Collection System}~\label{sec:motiontrans_data_collection} 
For human-robot cotraining, we need to collect both robot and human data~\cite{kareer2024egomimic}. For human data collection, we leverage a portable commercial VR device, which allows data to be collected anytime and anywhere. This provides great efficiency in gathering diverse motions and a wider range of tasks~\cite{hoque2025egodex}. For robot data collection, we use teleoperation to record demonstrations. The top-left side of Figure~\ref{fig:main_method} illustrates the two types of data collection systems.

\vspace{2mm}

\noindent \textbf{Human Data Collection with Portable VR Device.} 
We extend ARCap~\cite{chen2024arcap} to build our human data collection system (Figure~\ref{fig:main_hardware}(a)), incorporating a portable VR headset for recording hand keypoint positions \(K_t\), wrist poses \(W_t\) and camera poses, and an RGB camera for the image stream \(I_t\).

For hand pose recording, our goal is to capture the positions of hand keypoints \( K_t \) and human wrist poses \( W_t \) in the coordinate frame of the RGB camera (\( I_t \)). However, these information is recorded by VR device, placing it in the VR coordinate space. Therefore, we use a self-designed calibration method to transform all hand information from the VR coordinate space to the RGB camera, detailed in Appendix~\ref{app:camera_calibration}. For data collection, collectors are instructed to minimize head motion to approximate the static camera setting of real robot hardware, although slight movements are tolerated~\cite{qiu2025humanoidhuman}. To ensure data quality, we provide real-time feedback in the user’s VR view to guide collectors during data acquisition (Figure~\ref{fig:main_hardware}(b)). The feedback includes the RGB camera’s capture range and the positioning of the hands:
\begin{itemize}
    \item The range of images captured by the RGB camera is used to guide users to ensure their hands are always visible to the RGB camera~\cite{chen2024arcap}.
    \item The hand positioning tells collector whether the hand poses recorded by VR is strictly aligned with their hands in real time, thus provide information about the recording delay and accuracy. 
\end{itemize}

We also provide gesture interface to allow collector to abandon current recorded data anytime, if they think the data quality is not good enough considering the principles and feedback mentioned above. 

\vspace{2mm}

\noindent \textbf{Robot Data Collection with Teleoperation.} Since our goal is to achieve direct human-to-robot motion transfer, the robot hardware platform need to match the functionality of the human arm and hand. To this end, we choose the combination of a single robot arm and a dexterous robot hand as our hardware platform (Figure~\ref{fig:main_hardware}(c)). We develop our teleoperation system on Open-Television~\cite{cheng2024opentelevision}, which captures human wrist and hand poses in real time via a VR device and drives the robot to replicate these motions. Based on the collection system above, we collect our \textit{MotionTrans} human-robot datasets (Section~\ref{sec:experiment_setup} and Figure~\ref{fig:main_dataset}) for multi-task cotraining.

\subsection{Human Data Transformation to Robot Format}~\label{sec:motiontrans_data_processing}
As shown in the previous section, the raw human data collected from the VR device differs in format from robot data, which prevents it from being directly used for cotraining with robot policies~\cite{yang2025egovla, luo2025being}. To address this, we propose directly transforming human data into the robot's observation-action space~\cite{cheng2024opentelevision, lu2024mobile-teleop}. After transformation, the human data can serve as a form of ``supplementary robot data'' for training any mainstream end-to-end \textbf{robot} policy.

\vspace{2mm}

\begin{figure*}[t]
    \centering
    \includegraphics[width=1.0\linewidth]{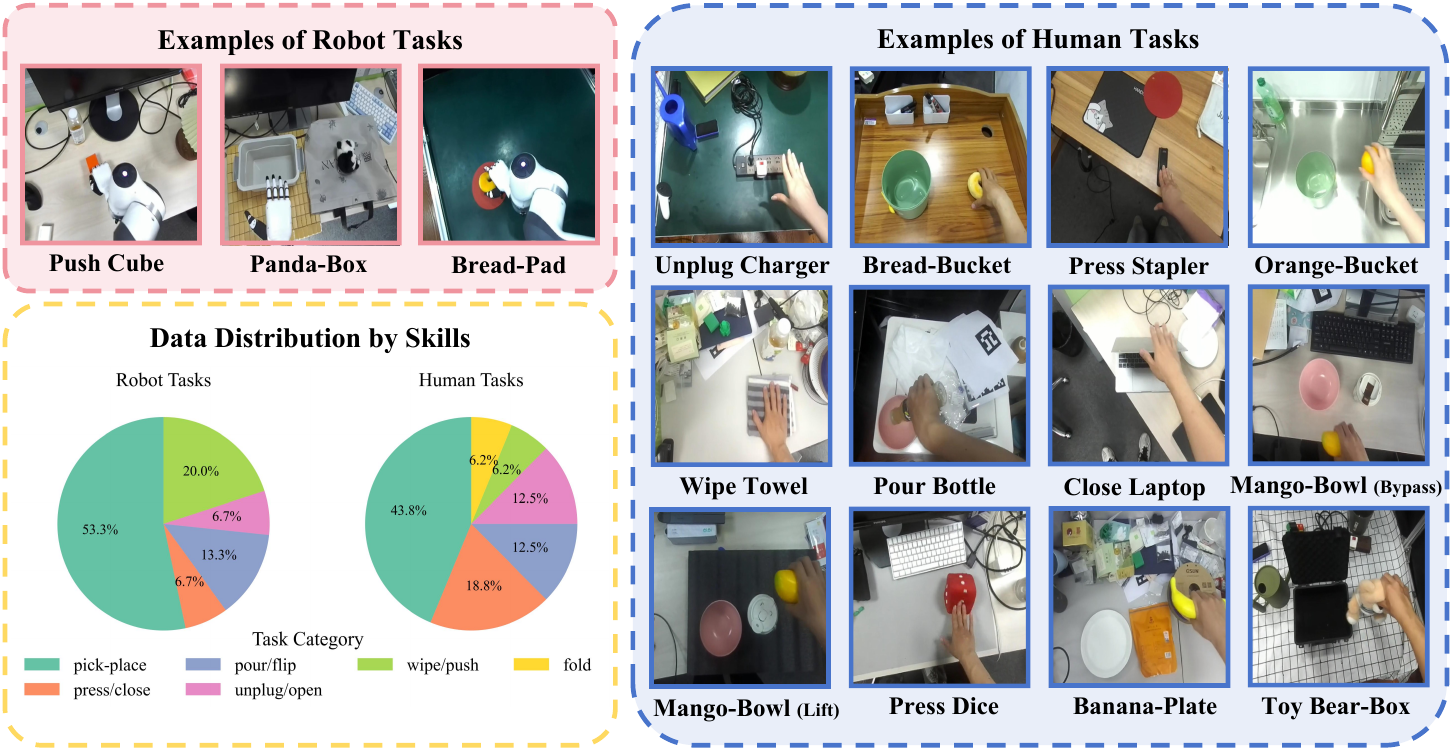}
    \caption{Illustration of the \textit{MotionTrans Dataset}, which comprises 3,213 demonstrations spanning 15 human tasks and 15 robot tasks collected across more than 10 scenes. The content in the () for the ``Mango-Bowl" task describes the method used to avoid obstacles. For statistical analysis, tasks are grouped by motion-similar skill categories. For human task ``Open Box+Panda-Box", it contains both open and pick-place skills. Detailed description and visualization of all 30 tasks are provided in Appendix~\ref{app:motiontrans_dataset}.}
    \label{fig:main_dataset}
\end{figure*}

\noindent \textbf{Transforming Observation-Action Space.} 
The observation-action space of the robot includes three components: image observation \( I_t \), proprioceptive state \( P_t \), and action \( A_t \) (refer to Section~\ref{sec:motiontrans_problem_define}). Both \( P_t \) and \( A_t \) are generated by stacking wrist poses $W_t$ and hand joint states $H_t$. Next, we describe the design for these components:

\begin{itemize}
    \item \textbf{Image observation $I_t$:} We use \textbf{egocentric} view for both human and robot data, as shown in Figure~\ref{fig:main_dataset}. The use of the similar image view makes the spatial relationships of objects in the scenes similar for accomplishing similar tasks, thus enabling similar motions to achieve those tasks.
    \item \textbf{Wrist poses $W_t$:} We use the \textbf{egocentric} camera coordinate system (camera captures \( I_t \)) for both human and robot data. This allows for the measurement of wrist poses in a unified coordinate system, ensuring that the spatial definitions of human and robot data are consistent. 
    \item \textbf{Hand joints state $H_t$:}  we employ the dex-retargeting library~\cite{qin2023anyteleop}, an optimization-based inverse kinematics solver, to map human hand keypoints $K_t$ to robot hand joint state $H_t$.
\end{itemize}
 
The design above converts human data into the same format as robot data, enabling us to directly replay human data on real-world robots. The replayable property of transformed human data proves how aligned our processed data is with robot data. 
By replaying human trajectories on a real-robot platform, we derive the following key observations: (O1) the speed of human manipulation is much faster than that of the robot, which affects safety and motion planning stability; (O2) there is a discrepancy between the distributions of human hand positions and the robot’s comfortable workspace (all defined in egocentric camera coordinate space). To alleviate these problem, we:
\begin{itemize}
    \item (O1) We \textbf{slow down human data by a factor of 2.25} via poses and hand joints state interpolation. More advanced techniques, such as the adaptive speed downsampling strategy~\cite{shi2025diversity}, are left for future exploration.
    \item (O2.1) We utilize \textbf{action-chunk-based relative poses}~\cite{chi2023diffusion, zhao2024aloha} as wrist action representation to reduce distribution mismatches between human and robot data. For instance, even if the robot's and human's hand positions differ in world space, their relative poses remain the same if they move forward at the same speed.
    \item (O2.2) We encourage collectors to \textbf{change viewpoints between trajectory recordings}. This enhances the diversity of positional relationships between the camera view and the targeted manipulation objects, thereby encouraging policies to adapt to a larger distribution of hand poses and, consequently, a larger workspace for the robot.
\end{itemize}

The methods and principles described above help reduce the gap between human and robot data, thereby improving the effectiveness of human-to-robot motion transfer. Prior works~\cite{lepert2025masquerade, lepert2025phantom, li2025h2r} have proposed rendering robots into human videos to further narrow the visual gap between the two domains. We replicate this rendering approach, as shown in Figure~\ref{fig:design_ablation_render_variant}, but did not observe significant improvements over directly training on human videos (Section~\ref{sec:discussion_design ablation}). Therefore, we do not employ this rendering technique by default in our framework.

\subsection{Weighted Multi-Task Human-Robot Cotraining}~\label{sec:motiontrans_cotraining}
By unifying the observation and action spaces, we enable joint training of human and robot data under a shared end-to-end robot policy. This section introduce the multi-task policy architectures we use and how we train these policies.

\vspace{2mm}

\noindent \textbf{End-to-End Multi-Task Policy Architectures.}  
We explore two popular end-to-end policy architectures:  
(1) \textbf{Diffusion Policy (DP)}~\cite{chi2023diffusion}: unlike the original single-task setup, we extend DP for multi-task training. Each task is associated with a learnable embedding, serving as a unique task condition. The visual encoder is replaced with DINOv2~\cite{oquab2023dinov2} to enhance visual perception ability~\cite{lin2024datascaling}.  
(2) \textbf{Vision-Language-Action model ($\pi_0$-VLA)}: we adopt network structure from~\cite{black2024pi_0}, a policy architecture integrating large-scale pretrained Vision-Language Models~\cite{steiner2024paligemma} for multimodal perception and instruction following. Since $\pi_0$-VLA supports language input, we directly use instructions to assign tasks. For $\pi_0$-VLA, we load $\pi_0$-droid pretrained checkpoints~\cite{pertsch2025fast} before training. 

\vspace{2mm}

\noindent \textbf{Unified Action Normalization.} To improve training stability, we apply Z-score normalization to both proprioceptive states and actions before training~\cite{chi2023diffusion, chi2024umi}. Previous human-robot cotraining works~\cite{kareer2024egomimic, tao2025dexwild} typically adopt independent normalization for human and robot data, arguing that it reduces the action gap between the two sources. However, in our motion-level evaluation setting, where the goal is to directly deploy human tasks on a real robot, this approach introduces a mismatch between training (human normalization) and inference (robot normalization), ultimately causing a performance drop (Section~\ref{sec:discussion_design ablation}). Therefore, we adopt unified action normalization across human and robot data within our framework.

\vspace{2mm}

\noindent \textbf{Weighted Human-Robot Cotraining.} Our final step is to design a strategy to train multi-task policies with the processed human-robot dataset. 
Given the potential imbalance between human and robot data~\cite{tao2025dexwild, qiu2025humanoidhuman}, we adopt a weighted cotraining strategy similar to~\cite{wei2025empirical}. The training objective over the combined dataset \( D = D_{\text{robot}} \cup D_{\text{human}} \) is defined as: $\mathcal{L}_{D} = \alpha \mathcal{L}_{D_{\text{robot}}} + (1-\alpha) \mathcal{L}_{D_{\text{human}}}$, where $\mathcal{L}$ denotes the loss function of imitation learning~\cite{chi2023diffusion, black2024pi_0}. In this paper, we set:
\[
\alpha = \frac{|D_{\text{human}}|}{|D_{\text{human}}| + |D_{\text{robot}}|}
\]
where $|D_{\text{robot}}|$ and $|D_{\text{human}}|$ representing the dataset sizes. This weight ensures that the sum of the weights for human and robot data is equal, leading to the balance of these two data sources. 
We also try domain adaptation training techniques like domain confusion~\cite{tzeng2017adversarial,tzeng2014mmd} to promote knowledge transfer from human domain to robot domain in our earlier exploration, but do not find it beneficial for motion transfer and it always leads to training instability. Thus, we choose the simplest weighted cotraining strategy in our framework. More details could be found in Appendix~\ref{app:domain_confusion}.

\begin{figure}[t]
    \centering
    \includegraphics[width=1.0\linewidth]{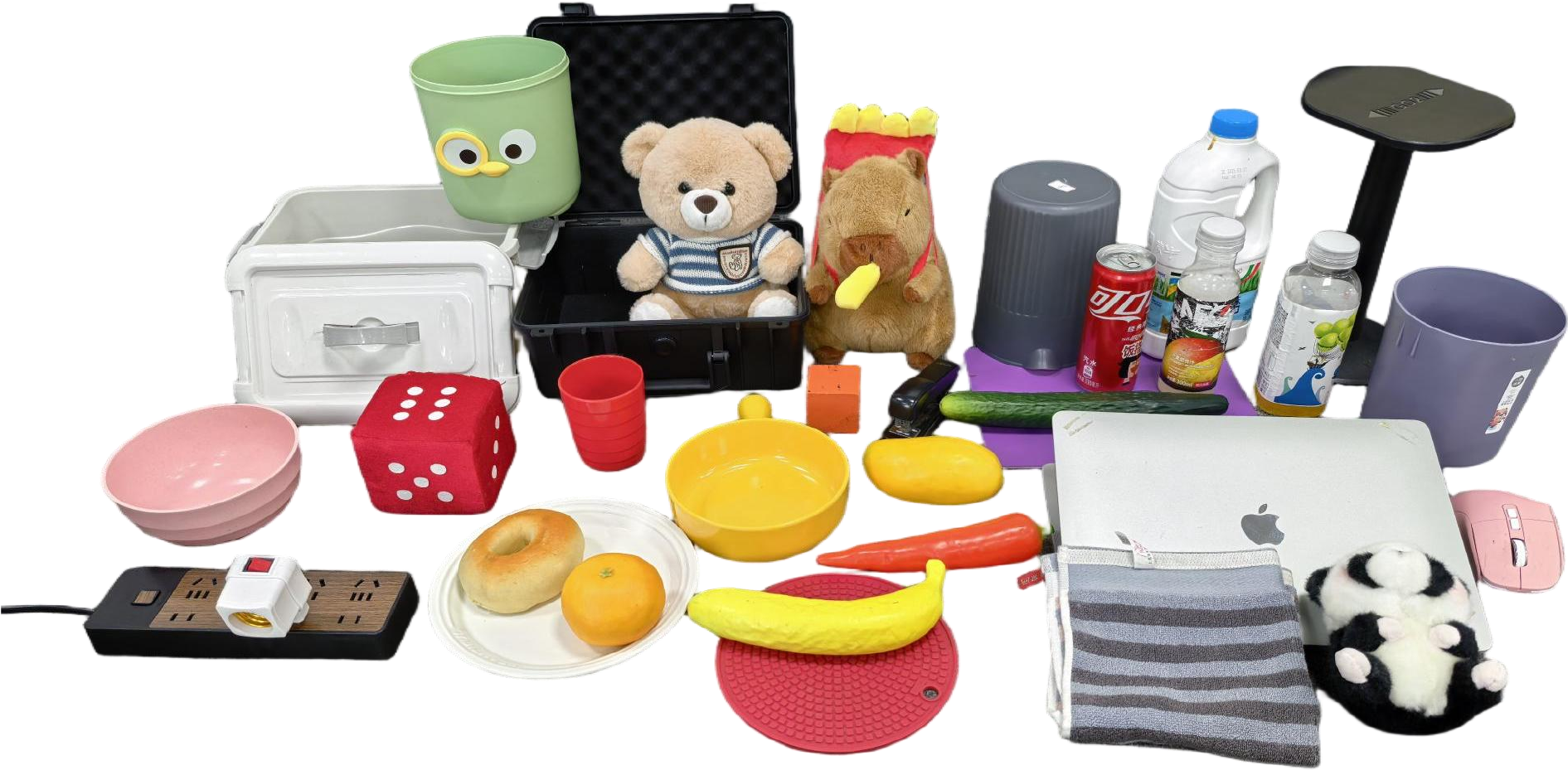}
    \caption{Illustration of all objects that have been manipulated for tasks in the \textit{MotionTrans Dataset.}}
    \label{fig:main_dataset_object}
\end{figure}

\section{Experiments}

In this section, we conduct experiments to verify the effectiveness of \textit{MotionTrans} for human-to-robot motion transfer. We first introduce our detailed experiment setup in Section~\ref{sec:experiment_setup}, including human-robot hardware platform, training datasets and evaluation tasks and metric. We then conduct experiments for both zero-shot (Section~\ref{sec:experiment_zero_shot}) and few-shot (Section~\ref{sec:experiment_few_shot}) settings, as demonstrated in Section~\ref{sec:motiontrans_problem_define}. Additionally, an ablation study on the key designs of \textit{MotionTrans} is performed (Section~\ref{sec:discussion_design ablation}). We also carry out experiments to explain the mechanism of human-to-robot motion transfer in Section~\ref{sec:discussion_mechanism_study}. The evaluation results of all robot tasks are shown in Appendix~\ref{sec:experiment_robot_task}. Finally, we verify the robustness of our results concerning visual backgrounds in Appendix~\ref{sec:discussion_visual_robustness}.

\subsection{Experiment Setup}~\label{sec:experiment_setup}

\vspace{-4mm}

\noindent \textbf{Hardware Platform.} For the robot hardware (Figure~\ref{fig:main_hardware}(c)), we use a Franka Emika robot arm~\cite{haddadin2022franka} in combination with a 6DoF Inspired Dexterous (Right) Hand~\cite{cheng2024opentelevision}. This combination mimics the functionality of a human right hand and arm. The robot is mounted on a movable lift table to facilitate data collection in various locations. A ZED2 camera is fixed to the table in an egocentric view to provide an image observation stream. The recorded images are first cropped to 640$\times$480 resolution and then resize to 224$\times$224. The VR device used for teleoperation is the Meta Quest 3~\cite{cheng2024opentelevision}. Calibration between the robot base and the robot perception camera is achieved through the DROID platform codebase~\cite{khazatsky2024droid}. 

For human data collection (Figure~\ref{fig:main_hardware}(a)), we use the Meta Quest 3 as our VR headset. To ensure consistency in image observations, we also employ a ZED2 camera to record RGB images and perform image cropping and resize, using the same setup as in the robot hardware platform~\cite{kareer2024egomimic}. The camera is attached to VR headset by a 3D-printed mounter~\cite{chen2024arcap}. The camera is connected to a laptop for data storage. The communication between the VR headset and the laptop is conducted via the local area network.

\vspace{2mm}

\begin{table}[t]
\centering
\resizebox{1.0\linewidth}{!}{%
\begin{tabular}{lcccccc}
\toprule
\multirow{2}{*}{Egocentric Datasets} & \multicolumn{3}{c}{Human} & \multicolumn{3}{c}{Robot} \\
\cmidrule(lr){2-4} \cmidrule(lr){5-7}
 & Demos & Tasks & In the Wild & Demos & Tasks & In the Wild \\
\midrule
EgoMimic~\cite{kareer2024egomimic}    & 2150  & 3$^\ddag$   & 0\%   & 1000  & 3   & 0\%   \\
PH$^{2}$D~\cite{qiu2025humanoidhuman}        & 26842 & 4$^\ddag$   & \textbf{100\%} & 1552  & 4   & 0\%   \\
\textit{MotionTrans} (Ours) & 1705  & \textbf{15}  & \textbf{100\%} & 1508  & \textbf{15}  &\textbf{ 50\%} \\
\bottomrule
\end{tabular}}
\caption{Comparisons of task-oriented egocentric human-robot cotraining datasets. $^\ddag$ indicates that the human and robot share the same task. For scene diversity (shown as the ``In the Wild” metric), we report the proportion of data collected in daily-life environments (in-the-wild setting~\cite{lin2024datascaling, tao2025dexwild}) rather than controlled lab settings. Our \textit{MotionTrans Dataset} demonstrates significant improvements in both task/motion coverage and scene diversity. These advances enable motion-level learning research that is not achievable with existing datasets.}
\label{tab:dataset_comparison}
\end{table}

\noindent \textbf{\textit{MotionTrans} Multi-Task Dataset.}
Here we introduce the \textit{MotionTrans Dataset}, which is used to train our policies. The dataset  contains 3,213 demonstrations across more than 10 scenes, covering 15 human tasks and 15 robot tasks.
A brief summary of the dataset is shown in Figure~\ref{fig:main_dataset}. The manipulated objects for all tasks are illustrated in Figure~\ref{fig:main_dataset_object}. 
The number of demonstrations for each human/robot task ranges from 40 to 150. 
The complete task list, the number of demonstrations and the visualizations for all 30 tasks are provided in Appendix~\ref{app:motiontrans_dataset}. A comparison with previous human-robot cotraining datasets is shown in Table~\ref{tab:dataset_comparison}.

For tasks, the human and robot task sets are non-overlapping. For motions, similar tasks across human and robot data (e.g., pick-and-place) share similar motion patterns but still exhibit notable differences. In addition, some motions appear only in the human dataset but not in the robot dataset, such as unplugging, closing, lifting, etc. Overall, the dataset covers a wide range of motions and skills, including pick-and-place, pouring, wiping, pushing, pressing, opening, etc. 
This wide coverage is proven crucial for successful motion transfer, as demonstrated in subsequent ablation studies (Section~\ref{sec:discussion_mechanism_study}).
For simplicity, we name pick-place task with ``pick object-place target" format, and name other task with ``verb noun" format in the main paper. For tasks with multiple steps, we name it as ``step1+step2" format.

To enhance the visual robustness of the policies~\cite{yuan2025roboengine} (Section~\ref{sec:discussion_visual_robustness}), such as robustness to different backgrounds and lighting conditions, we collect these data across various scenes~\cite{lin2024datascaling}. Each human task is collected in at least 4 different scenes. For robot tasks, about half of the data is collected in the \textit{``green table scenes"} (the scenes for the examples of the ``Bread-Pad" and ``Unplug Charger" task in Figure~\ref{fig:main_dataset}), with random disturbance objects placed on the table for approximately 80\% of the data. \textit{This scene is also designated as the default scene for our evaluation.} The other half of the robot tasks is collected in at least 4 scenes. To enrich language instructions for VLA training, we leverage GPT-4o~\cite{hurst2024gpt} to paraphrase and expand task descriptions in the dataset.

\vspace{2mm}

\begin{figure*}[t]
    \centering
    \includegraphics[width=1.0\linewidth]{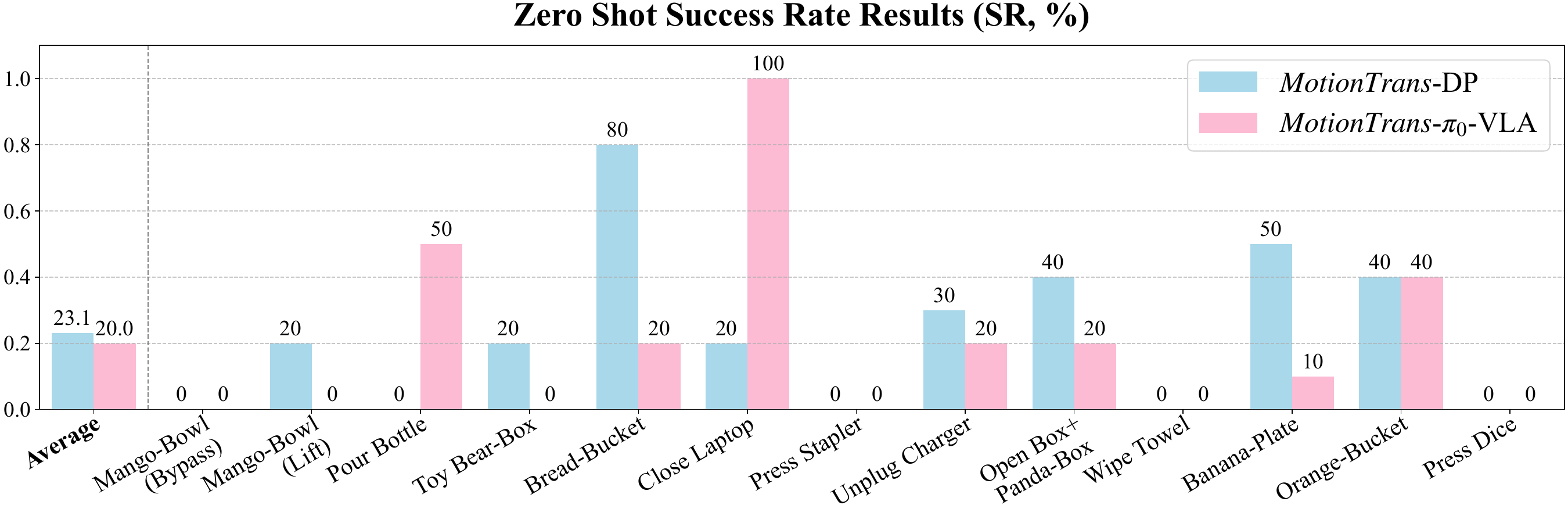}
    \includegraphics[width=1.0\linewidth]{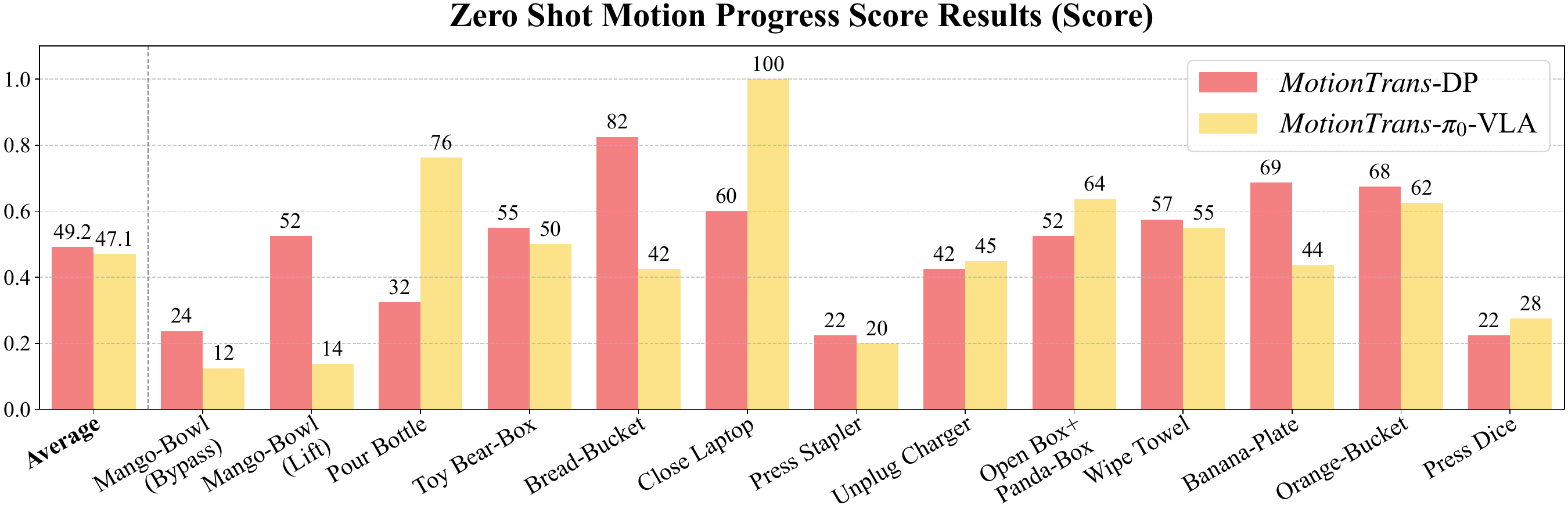}
    \caption{Results of \textit{MotionTrans} in the zero-shot experiment setting. The results show that both Diffusion Policy (DP)\cite{dp} and $\pi_0$-VLA\cite{black2024pi_0} achieve successful human-to-robot motion transfer. Even without any robot data for these human tasks, 9 tasks attain a non-zero success rate. For the remaining tasks, \textit{MotionTrans} still generates meaningful motion for task accomplishment, as indicated by a non-trivial Motion Progress Score.}
    \label{fig:zero-shot-result}
    \vspace{-4mm}
\end{figure*}

\noindent \textbf{Evaluation Tasks and Metrics.}
Since our goal is to understanding the effectiveness of human-to-robot motion transfer, 
we focus on \textbf{evaluating the performance of robot policies on the human tasks}. Among all 15 tasks in human dataset, there are two tasks (``Fold Towel" and ``Pour Milk Bottle") not been able to deploy to robot due to the hardware design limitation of robot hand (cannot be accomplished even if we use teleoperation). Therefore, we focus on discussing other 13 tasks in this research.
The list of all 13 evaluation tasks could be found in Figure~\ref{fig:zero-shot-result}.

We use the \textit{Success Rate (SR)} to evaluate the policy performance in accomplishing specific tasks. However, this metric alone is insufficient to reflect the effectiveness of motion transfer, as it ignores meaningful motion during task execution. For instance, a policy that demonstrates reaching for the target object should be rated higher than one that does not move at all. To address this limitation, we define a \textit{Motion Progress Score (Score)} to quantify the quality of policy motion for task completion. 
Detailed scoring rubrics for all tasks are provided in Appendix~\ref{app:score_rubics}. For clarity, we normalize the Score to a [0,1] range in the main paper.
For each task, we conduct 10 rollouts and calculate the average results for both metrics. We change \textbf{the object arrangement} for each rollout to cover a wide range of configurations of the task across the 10 rollouts.

\begin{figure*}[t]
    \centering
    \includegraphics[width=1.0\linewidth]{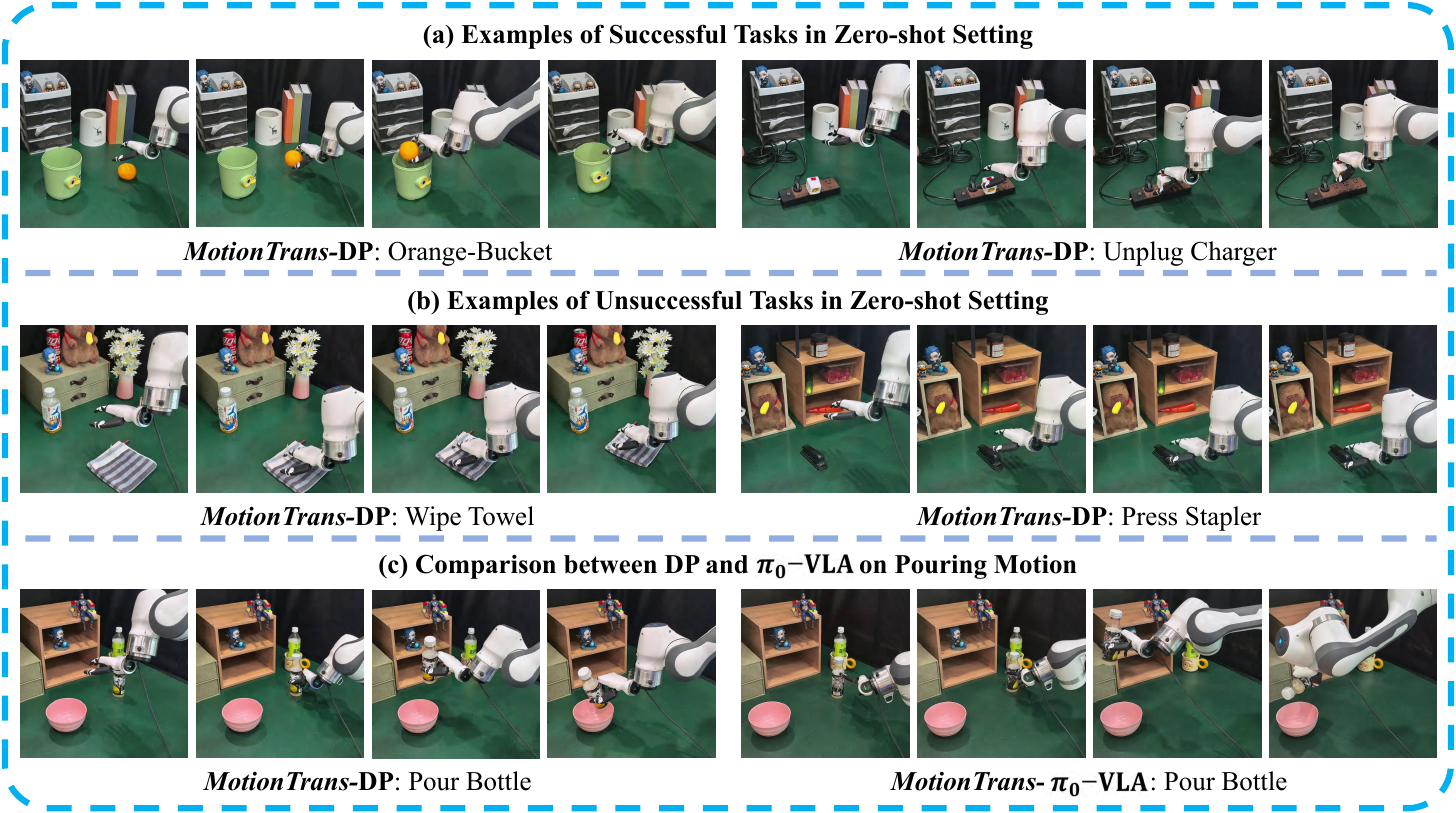}
    \caption{The visualizations for \textbf{zero-shot} human-to-robot motion transfer from our \textit{MotionTrans} framework. All tasks shown here do not involve any robot data collection and are learned from human data. These results demonstrate that the \textit{MotionTrans} enables explicit human-to-robot motion transfer for task completion through human-robot cotraining.}
    \label{fig:main_zero_shot_traj_vis}
    \vspace{-4mm}
\end{figure*}

\subsection{Zero-shot Experiment}~\label{sec:experiment_zero_shot}
The goal of the zero-shot experiment is to verify the effectiveness of direct human-to-robot motion transfer. 
We train policies using our \textit{MotionTrans Dataset}. Subsequently, we directly deploy policies to real robot hardware and evaluate the performance of tasks in human data. We refer to this as zero-shot setting because the policies learn motions from humans without any robot data collected for these human tasks.
We seek to answer the following questions:

\begin{itemize}
    \item (Q1.1) Can the policy directly learn to accomplish tasks in human data by human-robot cotraining, even without collecting any robot data for these tasks?
    \item (Q1.2) For tasks that cannot be accomplished, can the policy learn meaningful motion for task completion?
    \item (Q1.3) Is cotraining with robot data the key factor for achieving explicit motion transfer?
    \item (Q1.4) What is the difference in motion transfer effectiveness between different policy architectures?
\end{itemize}

\vspace{2mm}

\noindent \textbf{Experiment Details.} We train two end-to-end policies, Diffusion Policy (DP) and $\pi_0$-VLA (as mentioned in Section~\ref{sec:motiontrans_cotraining}). For DP, we train it for 300 epochs with a learning rate of \(5 \times 10^{-4}\) and 1024 batch size. For $\pi_0$-VLA, we train it for 160,000 steps with a learning rate of \(2.5 \times 10^{-5}\) and 192 batch size. Both models are trained with the AdamW optimizer~\cite{loshchilov2017adamw}. The training takes approximately 1.5 days and 2.5 days respectively. In this paper, we focus on enabling human-to-robot transfer for mainstream end-to-end policies. Therefore, we do not compare against zero-shot intermediate representation-based methods such as Vid2Robot~\cite{jain2024vid2robot}, General-Flow~\cite{yuan2024generalflow}, EgoZero~\cite{liu2025egozero}, ZeroMimic~\cite{shi2025zeromimic} etc., which are not compatible with such policies. Instead, our analysis centers on differences among end-to-end policy architectures (DP vs. $\pi_0$-VLA).

\vspace{2mm}

\noindent \textbf{(Q1.1) \textit{MotionTrans} enables policies to achieve non-trivial success rate across 9 tasks in the human dataset.} The results of the zero-shot experiment are shown in Figure~\ref{fig:zero-shot-result}. 
As shown in the results, 9 tasks achieve a non-trivial success rate. The average success rate on all 13 tasks is approximately 20\%. The visualization of two examples could be found in the Figure~\ref{fig:main_zero_shot_traj_vis}(a) (``Orange-Bucket" and ``Unplug Charger"). Among these tasks, pick-and-place tasks account for the vast majority. This can be attributed to (1) the simplicity of pick-and-place motion, (2) the similarity of motions between different pick-and-place tasks, and (3) the large number of such tasks in our dataset. Notably, for the cases where even if both the pick objects and place targets are not seen in robot tasks (e.g., the ``Orange-Bucket" task, visualized on the left side of Figure~\ref{fig:main_zero_shot_traj_vis}(a)), this type of task-level transfer is still possible. 

Other accomplished tasks include motions such as pouring, unplugging, lifting, opening and closing (pressing). While some tasks (e.g., ``Unplug Charger", 20\%) attain only limited success rate, the model consistently exhibit meaningful motion tendencies in unsuccessful rollouts, as will be discussed below. Overall, reaching the target emerged as the most reliable step across tasks, whereas precision-demanding actions such as grasping and infrequent motions in the dataset, such as unplugging, achieved limited success rates.


\vspace{2mm}

\noindent \textbf{(Q1.2) For unsuccessful tasks, \textit{MotionTrans} enables policies to learn meaningful motions toward task completion.} Figure~\ref{fig:zero-shot-result} shows that both DP and $\pi_0$-VLA achieve positive Motion Progress Scores across all tasks, with an overall average of about 0.5. This indicates that the policies are able to complete certain sub-processes for all evaluation tasks.  For instance, in the ``Wipe Towel" task, both DP and $\pi_0$-VLA learn the motion of ``push towel forward" to some extent (left side of Figure~\ref{fig:main_zero_shot_traj_vis}(b)). 
Moreover, we observe that human data enables the policy to identify spatial locations for almost all evaluated human tasks, which is represented as reaching the target manipulated objects (may only appear in human data) to some extent. An example of this is the ``Press Stapler" task in Figure~\ref{fig:main_zero_shot_traj_vis}(b): although the stapler is not seen in the robot data, the policy still performs approaching behavior.

\vspace{2mm}

\noindent \textbf{(Q1.3) Cotraining with robot data is the key factor for successful motion transfer.} We find that when robot data is not included for cotraining, the success rate across all tasks is 0\% for zero-shot setting. Generally, the policy trained solely on human data exhibits random motion when deployed on the robot. This demonstrates that cotraining with robot data is essential for explicit human-to-robot motion transfer, which could bridge the gap between humans and robots, allowing human motions to adapt to robot embodiment. 
A detailed analysis of the mechanism by which robot data support motion transfer can be found in Section~\ref{sec:discussion_mechanism_study}.

\vspace{2mm}
\begin{table}[t]
\centering
\renewcommand{\arraystretch}{1.1} 
\begin{tabular}{@{\hskip 6pt}l@{\hskip 20pt}c@{\hskip 20pt}c@{\hskip 6pt}} 
\toprule
                          & \textbf{\textit{MotionTrans}-DP} & \textbf{\textit{MotionTrans}-$\pi_0$-VLA} \\
\midrule
Toy Bear-Box              & 40                       & 0                           \\
Bread-Bucket              & 100                      & 20                          \\
Banana-Plate              & 50                       & 10                          \\
Orange-Bucket             & 70                       & 50                          \\
\cmidrule(lr){1-3}
\textit{Average}          & \textbf{65}             & \textbf{20}                \\
\bottomrule
\end{tabular}
\caption{The Success Rate of DP and $\pi_0$-VLA on all evaluation pick-and-place tasks for zero-shot setting. Generally, DP outperforms $\pi_0$-VLA during the grasping stage.}
\vspace{-4mm}
\label{tab:zero-shot-grasp}
\end{table}

\noindent \textbf{(Q1.4) DP and $\pi_0$-VLA each have their own advantages (manipulation precision and task adherence).} As shown in Figure~\ref{fig:zero-shot-result}, no single model excels across all tasks. On average, the performance of the two models is nearly identical. However, we observe that different models demonstrate their strengths on different tasks. Generally, DP performs better than $\pi_0$-VLA in precise manipulation stage, such as grasping, and exhibits stronger spatial location capabilities. An example of evidence for this is that, for all pick-and-place tasks, the average grasping success rate of $\pi_0$-VLA is 20\%, while DP achieves 65\% (Table~\ref{tab:zero-shot-grasp}).
In contrast, $\pi_0$-VLA shows stronger instruction following for motion generation in more cases. For example, in the ``Pour Bottle" task, we observed limited wrist rotation with DP, while $\pi_0$-VLA successfully performs the complete pouring action (Figure~\ref{fig:main_zero_shot_traj_vis}(c)). We hypothesize that this difference arises from a balance between visual perception and task semantic following. The model that focuses more on visual perception (DP) tends to achieve greater manipulation precision, whereas the model that emphasizes task semantics and instruction following ($\pi_0$-VLA) can adhere to task requirements more stringently.

\begin{figure*}[t]
    \centering
    \includegraphics[width=1.0\linewidth]{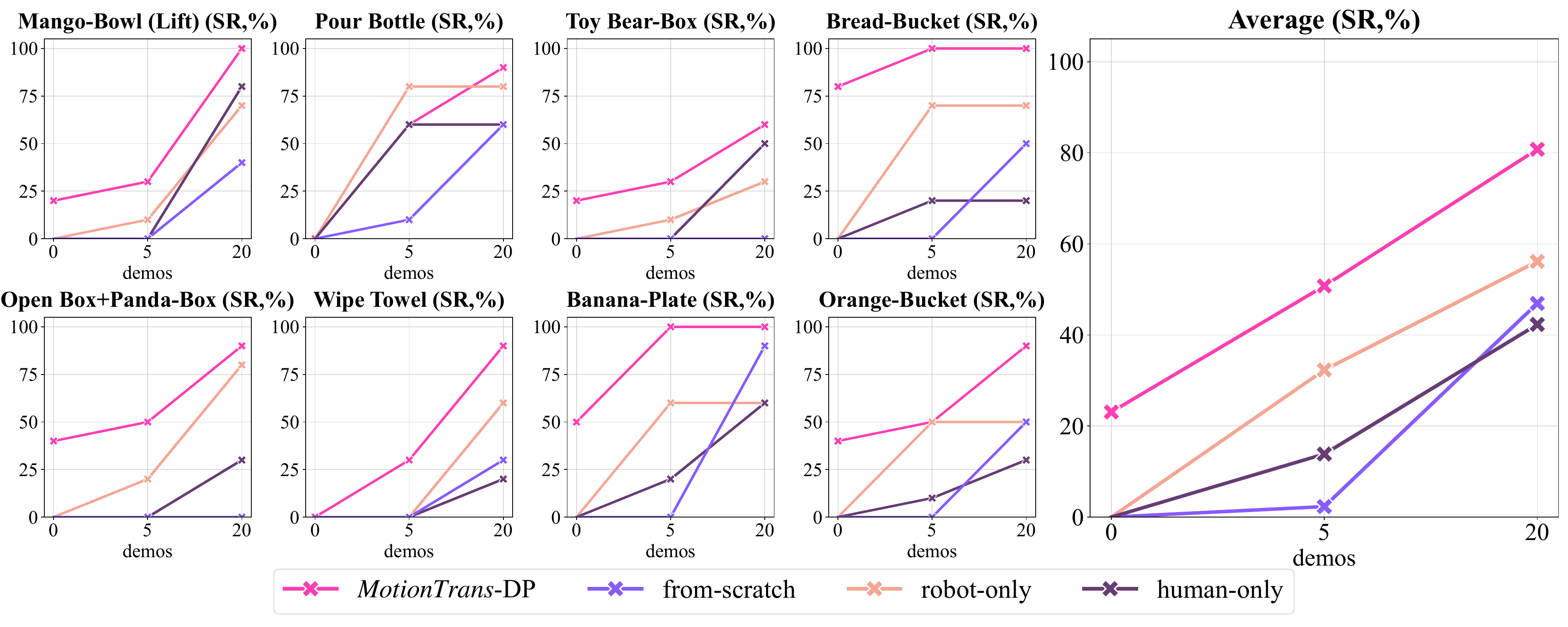}
    \caption{Results of the success rate for few-shot finetuning experiments. For readability, only the results of 8 example tasks are presented here. The Motion Progress Score results can be found in Appendix~\ref{app:score_fewshot_result}. From these results, we can conclude that both human and robot data during pretraining are important for improving finetuning performance.}
    \label{fig:few-shot-result}
    \vspace{-4mm}
\end{figure*}

\subsection{Few-shot Experiment}~\label{sec:experiment_few_shot}
In this section, we investigate whether motion transfer from human-robot cotraining can also enhance performance in a few-shot finetuning setting, where a limited number of robot demonstrations of human tasks are available for policy finetuning. We aim to answer the following questions:

\begin{itemize}
    \item (Q2.1) Will pretraining on \textit{MotionTrans Dataset} help improve policy finetuning performance?
    \item (Q2.2) What is the contribution of human data versus robot data for policy pretraining?
    \item (Q2.3) How does pretraining improvement vary with increasing finetuning data?
\end{itemize}

\vspace{2mm}

\noindent \textbf{Experiment Details.} Considering DP and $\pi_0$-VLA exhibit similar average performance in zero-shot experiments, we focus on DP architecture for computational resource efficiency in this part. We additionally collect 20 demonstrations for all human tasks in the default ``green table" evaluation scene, as mentioned in the dataset part in Section~\ref{sec:experiment_setup}. Subsequently, we perform 5-shot and 20-shot \textbf{multi-task finetuning}~\cite{bi2025hrdt} based on checkpoints previously trained on the \textit{MotionTrans Dataset}. We finetune DP with a learning rate of $1 \times 10^{-4}$ and a batch size of 256 for 200 epochs, employing the AdamW optimizer~\cite{loshchilov2017adamw}. The finetuning process requires 1 hour for the 5-shot setting and 4 hours for the 20-shot setting. 

We compared our method with three baselines to investigate the impact of different data components: (1) \textbf{``from-scratch"}, which means training policies without any pretraining; (2) \textbf{``robot-only"}, which entails pretraining solely on robot data from the \textit{MotionTrans Dataset} before finetuning; and (3) \textbf{``human-only"}, which is pretrained exclusively on human data.

\vspace{2mm}

\noindent \textbf{(Q2.1) Pretraining on \textit{MotionTrans Dataset} enable significant improvement for finetuning performance.} The success rate results of the few-shot experiments are presented in Figure~\ref{fig:few-shot-result}. The results of Motion Progress Score can be found in Appendix~\ref{app:score_fewshot_result}. We can see that policy pretrained on \textit{MotionTrans Dataset} gains around 40\% average success rate improvement compared to ``from-scratch" baseline. This is established for both 5-shot and 20-shot setting. These results prove that pretraining on human-robot data could provide useful motion prior~\cite{ye2023foundationrl,yang2025egovla} for downstream finetuning. 

\vspace{2mm}

\noindent \textbf{(Q2.2) Both robot and human data during pretraining are crucial for enhancing performance.} From Figure~\ref{fig:few-shot-result}, we can see that policy pretrained on both human and robot data (\textit{MotionTrans}) shows a significant advantage compared to human-only or robot-only pretraining. Besides, robot-only pretraining outperforms human-only pretraining on average. In our setting, robot pretraining uses data from the same embodiment but different tasks, whereas human pretraining uses data from the opposite case. We therefore conclude that maintaining the same embodiment in pretraining data is more important than exactly matching tasks. This is because the distribution of robot data is generally closer to the downstream robot finetuning distribution than human data, even when the tasks differ. Moreover, motions across different tasks often share similarities, so different robot tasks can still benefit downstream finetuning performance~\cite{black2024pi_0}.

\vspace{2mm}

\noindent \textbf{(Q2.3) Human-robot pretraining is more effective in low finetuning data region.} Finally, we analyze the impact of pretraining with varying amounts of finetuning data.
As shown in Figure~\ref{fig:few-shot-result}, the average performance of the policies improves consistently with an increase in finetuning data for all methods. 
However, the improvements are much larger in the 5-shot setting compared to the 20-shot setting. Moreover, when 20 finetuned demonstrations are available, the advantage of robot-only pretraining becomes minimal, and the benefit of human-only pretraining disappears. However, in the 5-shot setting, all pretraining methods show a significant advantage over the from-scratch baseline. 

\begin{table}[t]
\centering
\renewcommand{\arraystretch}{1.1} 
\begin{tabular}{@{\hskip 10pt}l@{\hskip 45pt}l@{\hskip 20pt}l@{\hskip 10pt}} 
\toprule
                    & \textbf{Score} & \textbf{SR (\%)} \\
\midrule
w/ Abs Pose         & 0.370 & \underline{10.0}    \\
w/ ED-Norm~\cite{kareer2024egomimic, tao2025dexwild}      & 0.341 & 8.4     \\
w/ Visual Rendering~\cite{lepert2025masquerade, li2025h2r}  & \underline{0.475} & \textbf{23.1}    \\
\cmidrule(lr){1-3}
\textit{MotionTrans}-DP & \textbf{0.492} & \textbf{23.1} \\
\bottomrule
\end{tabular}
\caption{Ablation results of design choices for \textit{MotionTrans}. The results are averages across all 13 evaluation human tasks. 
}
\label{tab:design_ablation}
\vspace{-4mm}
\end{table}

\begin{figure}[t]
    \centering
    \includegraphics[width=1.0\linewidth]{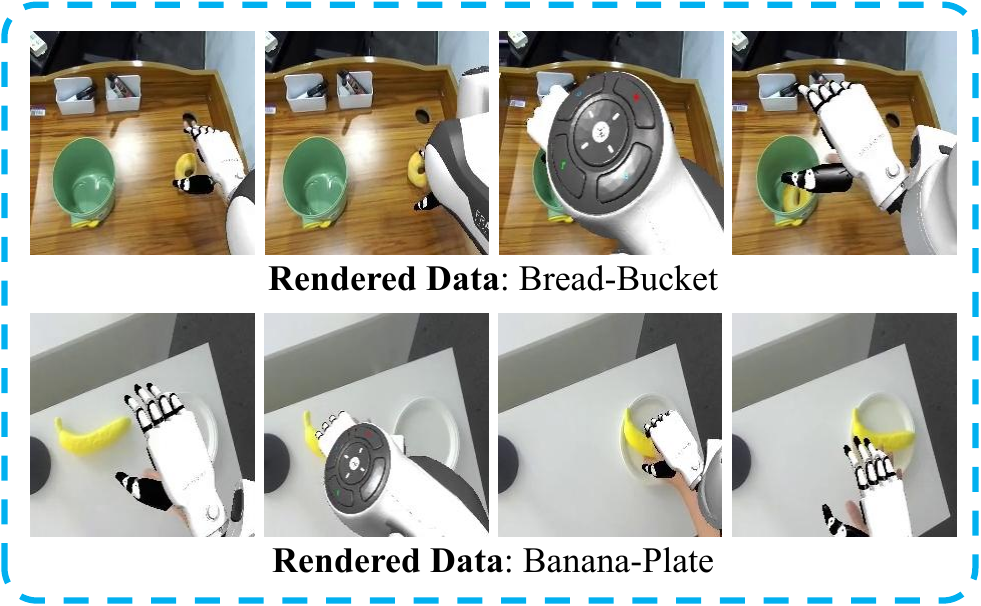}
    \caption{The visualizations of the rendered RGB observations for the \textbf{w/ Visual Rendering} variant in design ablation (Section~\ref{sec:discussion_design ablation}).}
    \label{fig:design_ablation_render_variant}
    \vspace{-4mm}
\end{figure}

\subsection{Design Ablation}~\label{sec:discussion_design ablation}
We conduct an ablation study on the key designs of \textit{MotionTrans}. 
We compare three variants of \textit{MotionTrans} in zero-shot setting experiments, including common techniques used in prior human-to-robot imitation learning: 

\begin{itemize}
    \item \textbf{w/ Abs Pose}: We replace the action-chunk-based relative pose~\cite{chi2024umi} with the original absolute egocentric pose for wrist action representation.
    \item \textbf{w/ ED-Norm}~\cite{kareer2024egomimic, tao2025dexwild}: We use independent action and proprioception normalization for human and robot data before policy training (\underline{E}mbodiment \underline{D}ependent \underline{Norm}alization).
    \item \textbf{w/ Visual Rendering}~\cite{lepert2025phantom, lepert2025masquerade, li2025h2r}: We first replay robot data in simulation, then crop the rendered robot and paste it to the original RGB image observation. 
    Visualizations of the rendered results are shown in Figure~\ref{fig:design_ablation_render_variant}.
\end{itemize}

\begin{table}[t]
\centering
\renewcommand{\arraystretch}{1.1} 
\begin{tabular}{@{\hskip 10pt}l@{\hskip 32pt}l@{\hskip 20pt}l@{\hskip 10pt}} 
\toprule
                    & \textbf{Score} & \textbf{SR (\%)} \\
\midrule
H-bucket         & 0.0 & 0    \\
H-bucket + R-pad      & 0.275 & 0     \\
H-bucket + R-platform  & 0.5 & 30    \\
H-bucket + R-pad + R-platform & 0.625 & 40 \\
\cmidrule(lr){1-3}
H-bucket + R-pad + R-platform + PP-set & 0.75 & 70 \\
all data (\textit{MotionTrans}) & 0.825 & 80 \\
\bottomrule
\end{tabular}
\caption{The results of the case study for the ``Bread-Bucket" task in zero-shot setting, including outcomes from training on different subsets of \textit{MotionTrans Datasets}. Detailed analysis could be found in Section~\ref{sec:discussion_mechanism_study}.}
\label{tab:mechanism_ablation}
\end{table}

\begin{figure}[t]
    \centering
    \includegraphics[width=0.95\linewidth]{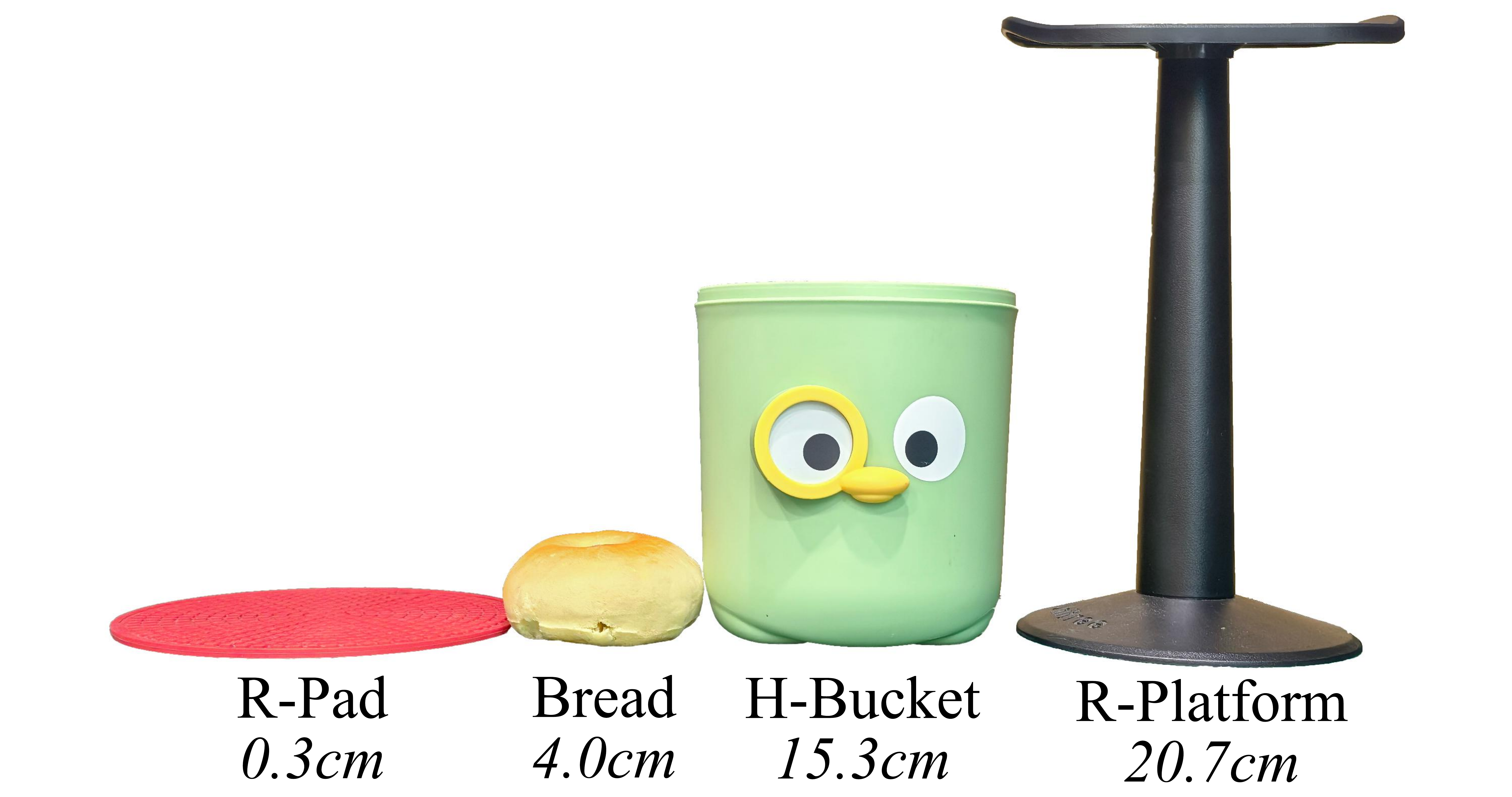}
    \caption{The visualizations of key objects used in the ``Bread-Bucket" case study are presented here. The height of each object is labeled beneath it.}
    \label{fig:ablation_task_object}
    \vspace{-4mm}
\end{figure}

\begin{figure*}[t]
    \centering
    \includegraphics[width=1.0\linewidth]{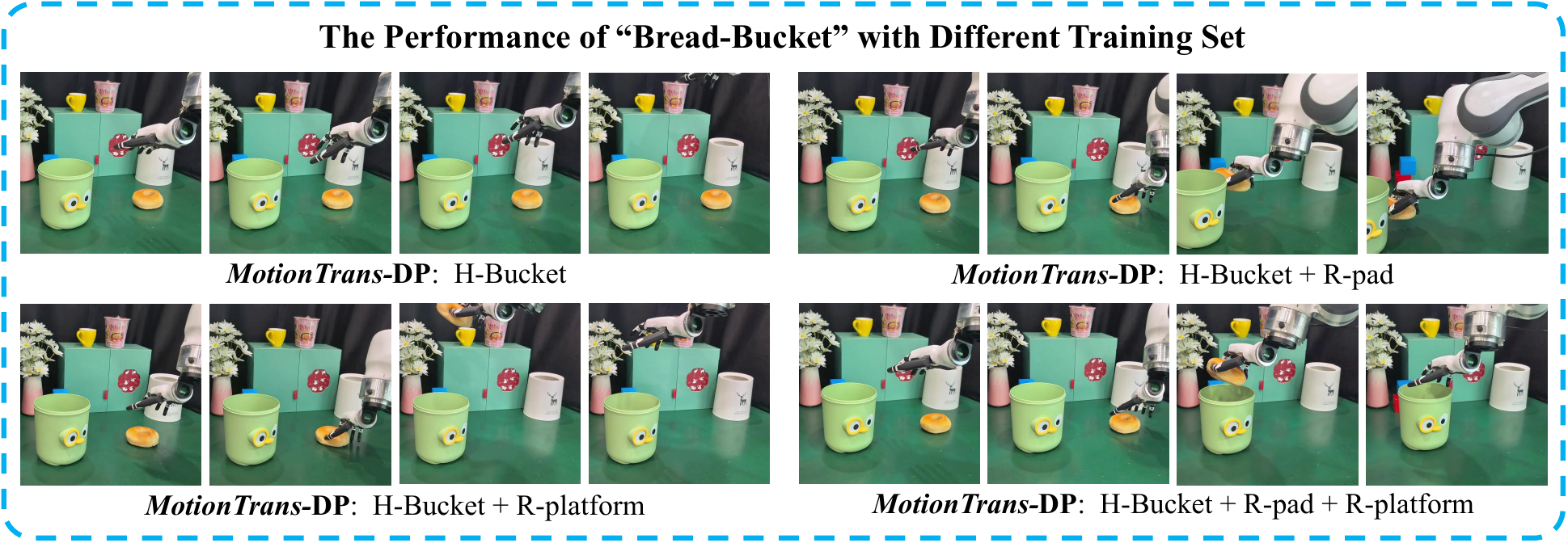}
    \caption{The visualizations of the \textit{MotionTrans}-DP results for the ``Bread-Bucket" task, trained con various combinations of human and robot tasks. By analyzing these results (Section~\ref{sec:discussion_mechanism_study}), we suggest that motion transfer occurs through the use of motion in human data to support robot motion interpolation for generating motions for these human tasks.}
    \label{fig:ablation_task}
    \vspace{-2mm}
\end{figure*}

The policy architecture chosen for all variants is Diffusion Policy (DP)~\cite{chi2023diffusion}. The results are averaged across all 13 evaluation tasks and shown in Table~\ref{tab:design_ablation}. We observe that \textbf{w/ Abs Pose} and \textbf{w/ ED-Norm} dramatically decrease the performance of human-to-robot motion transfer. For \textbf{w/ Abs Pose}, the usage of absolute pose increases the distribution difference between human and robot action label, prohibiting motion transfer, as discussed in Section~\ref{sec:motiontrans_data_processing}. For \textbf{w/ ED-Norm}, performance drops because the embodiment-dependent normalization creates a discrepancy in normalization between policy training and deployment. This contrasts with the phenomenon observed in visual robustness evaluations as demonstrated by previous works~\cite{kareer2024egomimic, tao2025dexwild}. When directly learning new motions and skills from human data, it is preferable to keep action normalization consistent between training and inference.

For \textbf{w/ Visual Rendering}, we find that performance is nearly the same as the non-rendered version. This may be explained by the fact that, despite appearing realistic to humans, the rendered results still include cues enabling neural networks to identify the embodiment domain. From this perspective, they offer little distinction from the original human videos. One potential solution is to also conduct inpainting during policy inference~\cite{lepert2025phantom}, but may lead to additional computational overhead and policy delay. All these results demonstrate that, when considering \textit{motion-level} transfer and evaluation, the effectiveness of certain designs may differ from their effectiveness when using human data to improve visual robustness~\cite{tao2025dexwild} or the training efficiency~\cite{kareer2024egomimic} for in-domain robot tasks.

\begin{figure}[t]
    \centering
    \includegraphics[width=1.0\linewidth]{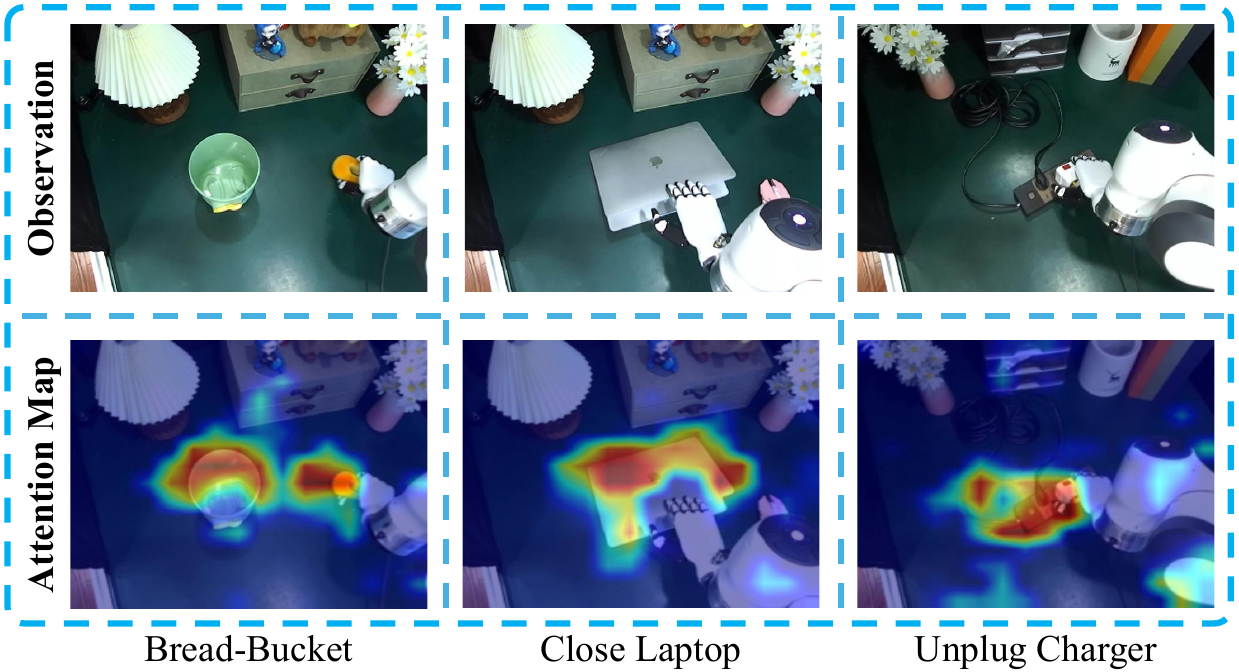}
    \caption{The visualization of the attention map from the DINO encoder~\cite{oquab2023dinov2} for \textit{MotionTrans}-DP, based on the Grad-Cam toolkit~\cite{selvaraju2017grad}. This shows that the vision encoder learns to focus on the target manipulation objects for tasks in human datasets, even when the embodiment changes to a robot during inference.}
    \label{fig:dino_vis_examples}
    \vspace{-4mm}
\end{figure}

\subsection{Analysis of Motion Transfer Mechanisms}~\label{sec:discussion_mechanism_study}
We have verified the feasibility of explicit human-to-robot motion transfer in our zero-shot experiments (Section~\ref{sec:experiment_zero_shot}). However, the underlying mechanisms of this transfer remain underexplored. In this section, we design experiments to investigate these mechanisms from three perspectives: (1) how actions transfer, (2) how visual perception transfers, and (3) the scaling trends of motion transfer. We then describe the experimental setup and present the corresponding conclusions.

\vspace{2mm}

\noindent \textbf{(Q3.1) How Do Actions Transfer?}  
To answer the question, we conduct a case study by down-sampling the number of tasks. We train policies on different subsets of \textit{MotionTrans Dataset} and compare their performance, gaining insights into how varying training tasks and motions influence the actions generated during real-robot inference. 
We select the task ``Bread-Bucket" as the evaluation task for our case study, as it already demonstrates a high success rate (80\%) in zero-shot settings, indicating effective motion transfer. Since ``action’’ is an abstract concept, we focus on a concrete dimension: \textbf{the height of object placement}. Three tasks with varying placement heights for the ``bread’’ object are selected to create training subsets:

\begin{itemize}
    \item \textbf{(Human Data) Bread-Bucket:} evaluation task, denoted as ``H-bucket".
    \item \textbf{(Robot Data) Bread-Pad:} placing bread on a thin red pad, ``R-pad".
    \item \textbf{(Robot Data) Bread-Platform:} placing bread on a tall black platform, ``R-platform".
\end{itemize}

Visualizations and placement heights of objects in these tasks are shown in Figure~\ref{fig:ablation_task_object}, and evaluation results in Table~\ref{tab:mechanism_ablation} (rows 1–4). Trajectory visualizations are provided in Figure~\ref{fig:ablation_task}. 
Results show that training only on human data tends to cause ambiguity during deployment on robots, consistent with our zero-shot findings (Section~\ref{sec:experiment_zero_shot}). Cotraining with a single robot task appears to bias the policy toward that specific placement height. When both R-pad and R-platform are included, the policy shows evidence of interpolating across placement heights, leading to bucket-height-aware motions.

Based on the results, \textbf{we hypothesize that: the actions for human task completion during real-robot inference are generated by interpolating actions from robot data} (e.g., from R-Pad at 0.3cm and R-Platform at 20.7cm to H-Bucket at 15.3cm). This interpolation ability is learned by training on task-aware motions in human data. When the policy encounters the robot embodiment during inference, it still generates actions within the safe manipulation range as defined by the robot data. However, \textbf{task-specific elements}, such as task identifiers or task-related objects in image observation, trigger the policy to activate the interpolation process to generate task-aware motion.

\vspace{2mm}

\noindent \textbf{(Q3.2) How Do Visual Perception Transfers?}  
We next examine visual perception. To understand the impact on policy visual perception when the embodiment changes from human during training to robot during inference, we visualize attention maps of trajectories from zero-shot \textit{MotionTrans}-DP policy using the DINOv2 encoder~\cite{oquab2023dinov2} and Grad-CAM~\cite{selvaraju2017grad}. 

Example results are shown in Figure~\ref{fig:dino_vis_examples}. The findings indicate that \textbf{the visual encoder attends to target objects in tasks of human data}, despite embodiment shifts at deployment (from human to robot). This embodiment-invariant, task-aware representation allows the policy to generate task-relevant motions during robot deployment and explains its ability to locate target objects even if they appeared only in human data.

\vspace{2mm}

\noindent \textbf{(Q3.3) Is There a Scaling Trend in Motion Transfer?}  
Finally, we study the effect of task diversity and motion coverage. we hypothesize that a wider range of motion and task coverage may enhance the policy's ability for motion interpolation and visual attention as mentioned before, thus leading to improved transfer performance. We verify this through subset training comparison experiment, similar to Q3.1. We introduce a new task subset ``PP-set" compared to Q3.1, which includes data from two robot tasks ``Mango-Bowl" and ``Capybara-PPad" and two human tasks ``Banana-Plate" and ``Toy Bear-Box". 

The performance of \textit{MotionTrans}-DP trained on different subsets for the ``Bread-Bucket" task is shown in the last 3 rows of Table~\ref{tab:mechanism_ablation}.
The results indicate a steady improvement with increased task coverage, suggesting that \textbf{motion transfer may benefit from broader task-related motion coverage}. While our study is based on a limited subset, these findings provide preliminary evidence of a potential scaling trend in human-to-robot motion transfer.

\subsection{Supplementary Experiment Results}

The evaluation results of all tasks in robot data are shown in Appendix~\ref{sec:experiment_robot_task}. We also verify the robustness of our results concerning visual backgrounds in Appendix~\ref{sec:discussion_visual_robustness}.

\section{Conclusion}\label{sec:conclusion}

In this paper, we propose \textit{MotionTrans}, a framework that achieves motion-level learning from human data for end-to-end robot policies. The experiments show that our method achieves explicit human-to-robot motion transfer in a zero-shot setting and significantly improves finetuning performance in a few-shot setting. We identify two key factors for successful motion transfer: (1) cotraining with robot data, and (2) broader coverage of motions and tasks, which leads to better transfer performance. We hope that the new motion-centric insights that we propose could enhance the utilization of human data in robot policy learning in more effective ways.

\vspace{2mm}

\noindent \textbf{Limitations and Future Directions.} Our largest limitation is that the height perception ability of the policies is still limited, which causes them to sometimes fail to reach the correct height when considering in-the-wild scenes. This limitation arises from our monocular egocentric perception setting, which may be addressed by adding wrist camera for both human and robot hardware platforms~\cite{xu2025dexumi, tao2025dexwild}. Another limitation is that our study is still limited to self-collected human dataset. Extending motion-level learning to larger, internet-scale datasets, as in~\cite{luo2025being}, is left for future work.

\section*{Acknowledgments}

This work is supported by the National Key R\&D Program of China (2022ZD0161700), National Natural Science Foundation of China (62176135, 62476011), Shanghai Qi Zhi Institute Innovation Program SQZ202306, the Tsinghua University Dushi Program, the grant of National Natural Science Foundation of China (NSFC) 12201341.

We would like to express our sincere gratitude to Shuo Wang, Gu Zhang, Enshen Zhou, Haoxu Huang, Jialei Huang, Ruiqian Nai, Zhengrong Xue, Junmin Zhao, and Weirui Ye for their valuable discussions. We are especially grateful to Ruiqian Nai and Fanqi Lin for their assistance with the implementation of $\pi_0$-VLA, and to Yankai Fu for his support with the hardware implementation. Our thanks also extend to the SpiritAI and InspireRobot team for their assistance.

\bibliographystyle{plainnat}
\bibliography{references}

\newpage

\

\newpage

\section{APPENDIX}

\subsection{Details of \textit{MotionTrans Dataset} and All Tasks}\label{app:motiontrans_dataset}
\begin{figure*}[h]
    \centering
    \includegraphics[width=0.85\linewidth]{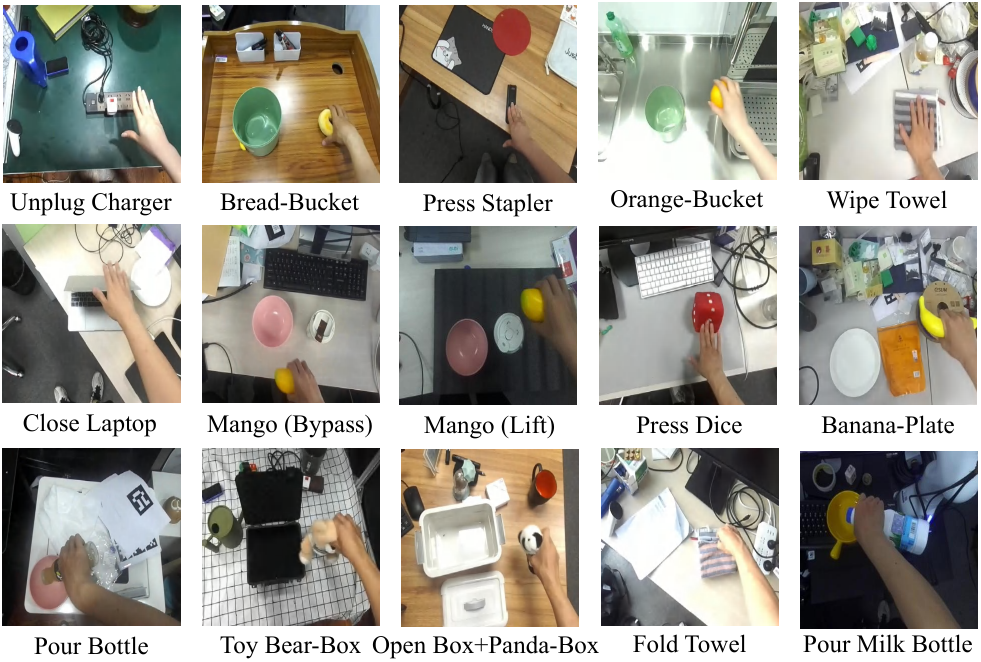}
    \caption{The visualizations of all 15 human tasks in the egocentric view.}
    \label{fig:app_task_human}
\end{figure*}

\begin{figure*}[h]
    \centering
    \includegraphics[width=0.85\linewidth]{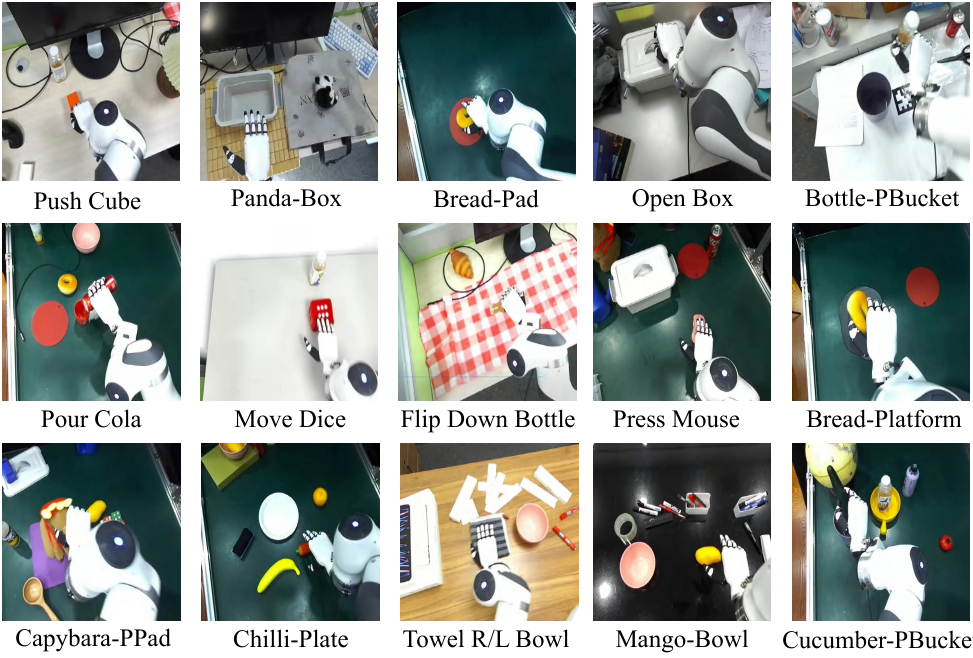}
    \caption{The visualizations of all 15 robot tasks in the egocentric view.}
    \label{fig:app_task_robot}
\end{figure*}

Here we present the details of all tasks in \textit{MotionTrans Dataset}. 
The visualization and descriptions / VLA-prompt of all 15 human tasks could be found in Figure~\ref{fig:app_task_human} and Table~\ref{tab:app_human_tasks}. All 15 robot tasks could be found in could be found in Figure~\ref{fig:app_task_robot} and Table~\ref{tab:app_robot_tasks}.

\subsection{Rubrics of Motion Progress Score}\label{app:score_rubics}
Table~\ref{tab:motion_progress} provides the detailed rubrics for our Motion Progress Score metric. The scores are allocated to the different motions / stages required to complete the task, with a maximum score of 8 points.

\subsection{Calibration between VR Headset and RGB Camera}\label{app:camera_calibration}
\begin{figure*}[t]
    \centering
    \includegraphics[width=1.0\linewidth]{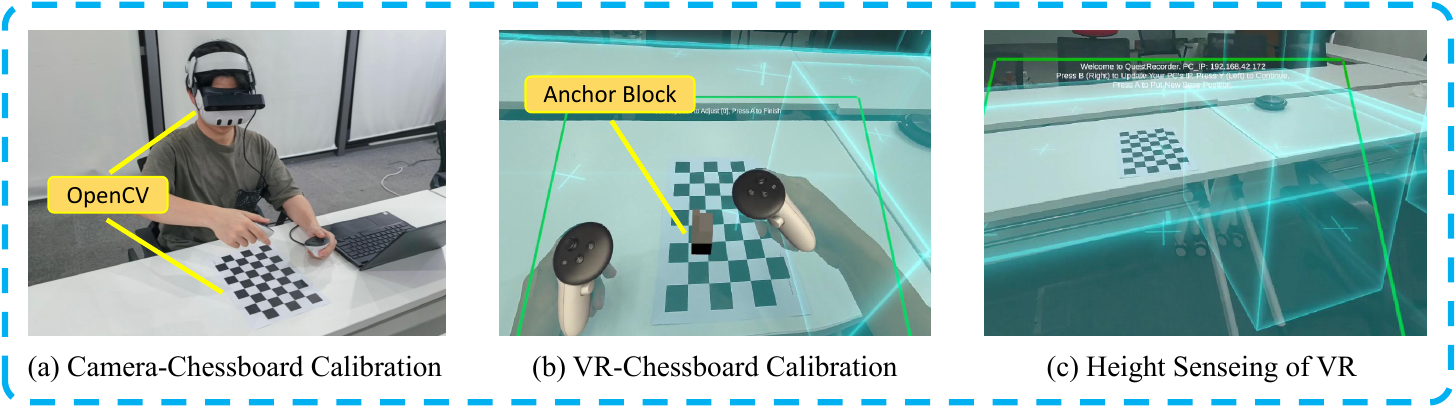}
    \caption{The illustration of the calibration process used to transform data from the VR coordinate space to the RGB camera space. Detailed demonstration can be found in Appendix~\ref{app:camera_calibration}.}
    \label{fig:app_calibration}
\end{figure*}

In human data collection (Section~\ref{sec:motiontrans_data_collection}), our goal is to record hand pose information captured by a VR device in the RGB camera's coordinate system.
To transform hand poses from the VR coordinate space to the RGB camera, we need to solve the transformation between the two cameras. We achieve this by applying a chain-style calibration.
We place an ArUco calibration chessboard~\cite{an2018charuco} on the table and ask users to sit facing it without moving their heads, as illustrated in Figure~\ref{fig:app_calibration}. We then perform two calibrations: 

\begin{itemize}
    \item \textbf{Camera-Chessboard Calibration} (Figure~\ref{fig:app_calibration}(a)). Solve \( T_{cam} \), the pose of the RGB camera based on the chessboard coordinate, using the vision-based calibration method~\cite{an2018charuco} (OpenCV library~\cite{bradski2000opencv}).
    \item \textbf{VR-Chessboard Calibration} (Figure~\ref{fig:app_calibration}(b)). Solve \( T_{vr} \), the pose of the VR camera based on the chessboard coordinate by asking the user to place an anchor block on the origin of the chessboard coordinate. We then directly read the coordinate of the anchor block (i.e., the origin) in the VR camera's coordinate space using the VR app, thus obtaining \( T_{vr} \) by inverting the reading result. To improve placement accuracy, we use the depth sensing built into the VR headset to fit the desktop height (the blue plane in Figure~\ref{fig:app_calibration}(c)), allowing users to only adjust the anchor block's position forward, backward, left, and right.
\end{itemize}

By using the chessboard as the bridge, the transformation used to convert hand poses from the VR to the RGB camera coordinate can be expressed as \( T_{cam}^{-1} T_{vr} \).

\begin{figure}[t]
    \centering
    \includegraphics[width=1.0\linewidth]{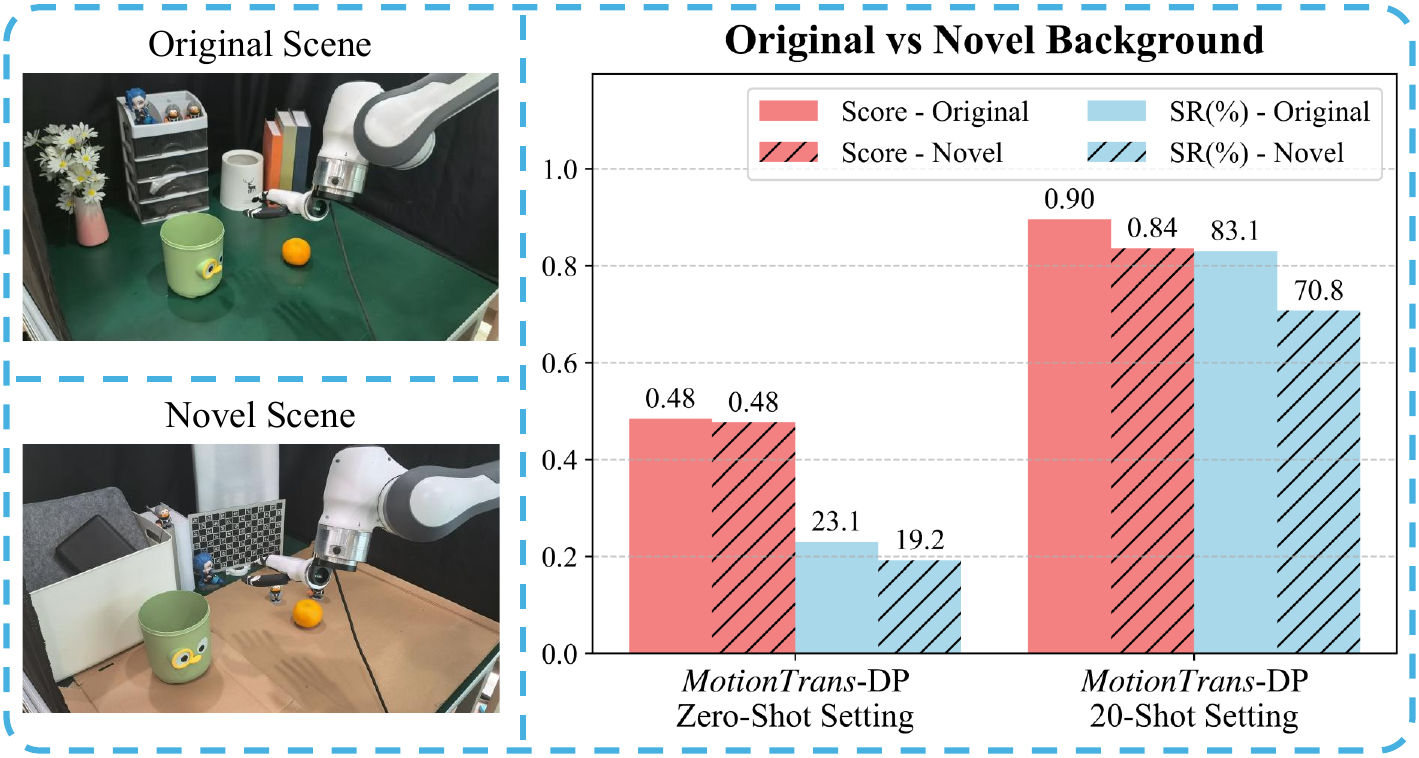}
    \caption{Illustration of the visual background robustness experiment and results. All results are averaged across all 13 evaluation human tasks. For the novel background, the performance drops slightly but remains at a persuasive level. This prove the robustness of our motion transfer results.}
    \label{fig:main_robustness}
\end{figure}

\subsection{Policies Implementation Details.}\label{app:policy_implementation}
For the robot observation-action space (Section~\ref{sec:motiontrans_problem_define}), we set the proprioception history $T_p=2$ and the action horizon $T_A=16$. The representation of the rotation component of wrist poses is chosen as the first two rows of the rotation matrix, as demonstrated in \cite{chi2023diffusion}. For policy control, we use 10 fps for both data collection and policy inference. For Diffusion Policy (DP) backbone, the task-embedding dimension is set as 16. The proprioception state is encoded via a 4-layer MLP. The DINOv2 vision encoder utilizes DINOv2-base pretrained checkpoints~\cite{oquab2023dinov2}, and during training, we unfreeze the weights of the DINOv2-base encoder. We first concatenate the task embedding with the features from the vision and proprioception encoder, and then input the concatenated features into the U-Net-based Diffusion head for action generation~\cite{chi2023diffusion}.

\subsection{Domain Confusion Training Framework}\label{app:domain_confusion}
We also tried the domain confusion framework~\cite{tzeng2017adversarial, tzeng2014mmd} in our earlier exploration. The key idea is to: (1) train a classifier to identify the embodiment from the features generated by the policy encoder.
(2) The policy's target is to generate embodiment-invariant features that mislead the classifier. These embodiment-invariant features thus facilitate better embodiment-agnostic knowledge transfer.
(3) Train these two models in an adversarial manner, allowing them to improve themselves by competing against each other.

Following the domain adaptation framework, we additionally train a binary classifier $C$ to classify whether a data point is from the human or robot domain. The input of $C$ is the concatenation of image features and proprioception features generated by policy $P$, as demonstrated in Appendix~\ref{app:policy_implementation}. Since we only want $C$ to classify based on embodiment/domain, rather than depend on shortcuts like task-specific content in $D_{\text{human}}$ / $D_{\text{robot}}$ (e.g., task-related objects in image observations), we trained $C$ on an augmented version of $D_{\text{human}}$ / $D_{\text{robot}}$: we first used GroundingDINO~\cite{liu2024grounding} / RoboSAM~\cite{yuan2025roboengine} to segment the robot out and then pasted it onto a novel MIL-texture background~\cite{finn2017mil, yuan2025roboengine}. We represent the augmented version as $D_{\text{human}}^{aug}$ / $D_{\text{robot}}^{aug}$. The final training loss $\mathcal{L}$ of policy $P$ can be represented as:
\begin{align}
&\mathcal{L}_{dc} = D_{KL}(C_{frozen}(P_{encoder}(s))\ ||\ U) \\
&\mathcal{L} = \mathcal{L}_{D} + \alpha \mathcal{L}_{dc}
\end{align}
where $\mathcal{L}_D$ is the imitation learning loss described in Section~\ref{sec:motiontrans_cotraining}, $P_{encoder}$ is the image and proprioception encoder of policy $P$, $\mathcal{L}_{dc}$ is the domain confusion loss (~\cite{tzeng2017adversarial}), $U$ is a binary uniform distribution, and $s$ is the data points randomly sampled from $D_{\text{human}}^{aug}$ / $D_{\text{robot}}^{aug}$. The $\mathcal{L}_{dc}$ encourages policy $P$ to generate similar features when only considering embodiment differences, thereby leading to embodiment-invariant features that enhance human-to-robot knowledge transfer.
We train policy $P$ (based on $\mathcal{L}$) and classifier $C$ (based on binary cross entropy loss, BCELoss) in an adversarial manner, which means that we iteratively train these two models. When training one model, we freeze the weight of another model. More details can be found in \cite{tzeng2017adversarial}. The architecture of $C$ is the same as that of $P$, except for changing the final MLP to a classifier head. 

\begin{figure*}[t]
    \centering
    \includegraphics[width=1.0\linewidth]{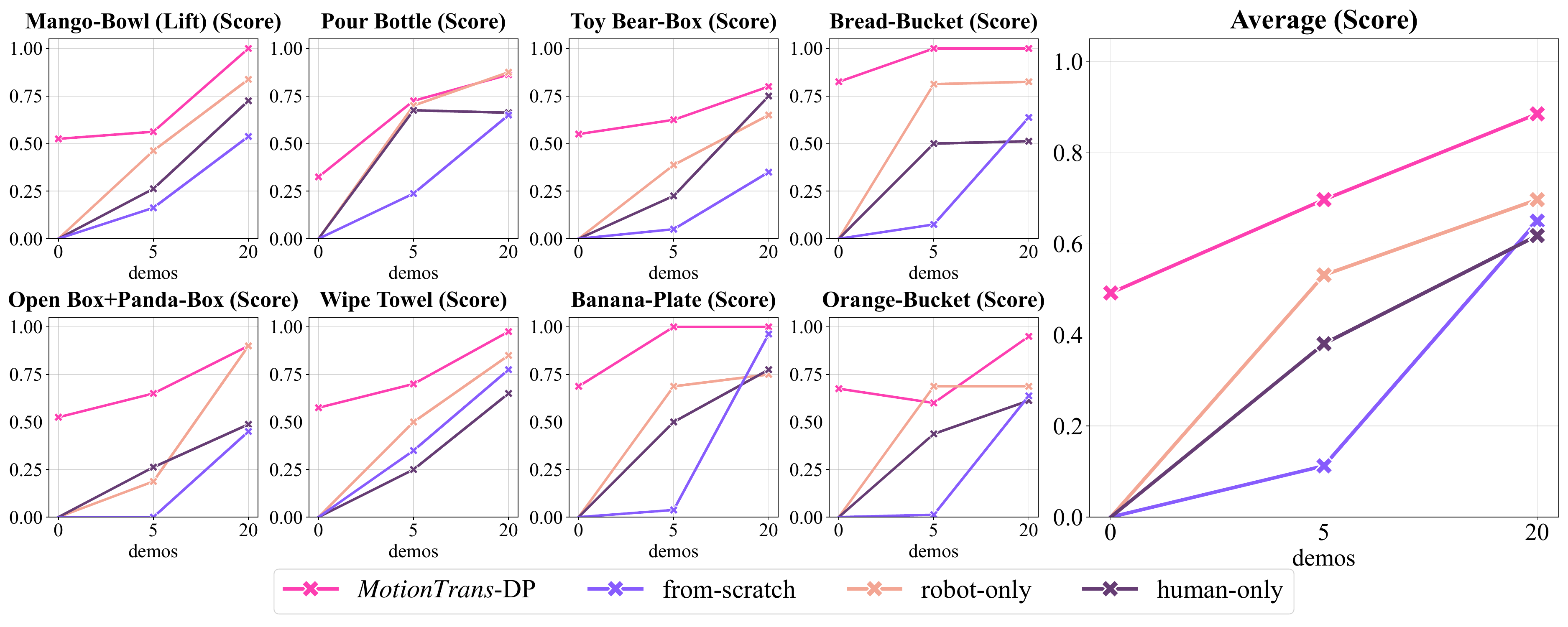}
    \caption{Results of the Motion Progress Score for few-shot finetuning experiments.}
    \label{fig:few-shot-result-score}
\end{figure*}

However, although we have tried our best to tune different settings, we still find that this kind of adversarial training tends to lead to mode collapse and training instability~\cite{arjovsky2017wasserstein}. This is reflected in the sudden jumps in the value of a certain loss during the training process, as well as the robot's inability to perform meaningful actions during downstream policy deployment. 
The key insight we gained from this experience is that rather than only relying on updating the algorithm or model, improving the scale and quality of data may be a more straightforward way to enhance transfer effectiveness. With enough task-related motion coverage, the simplest weighted cotraining framework shows the strongest transfer performance in our setting.

\subsection{Few-shot Results of Motion Progress Score}\label{app:score_fewshot_result}
The results of Motion Progress Score for few-shot experiment (Section~\ref{sec:experiment_few_shot}) are shown in Figure~\ref{fig:few-shot-result-score}. The conclusion drawn from the Motion Progress Score aligns with that from the Success Rate (Section~\ref{sec:experiment_few_shot}).

\subsection{Robot Tasks Experiment}~\label{sec:experiment_robot_task}
\begin{figure*}[t]
    \centering
    \includegraphics[width=1.0\linewidth]{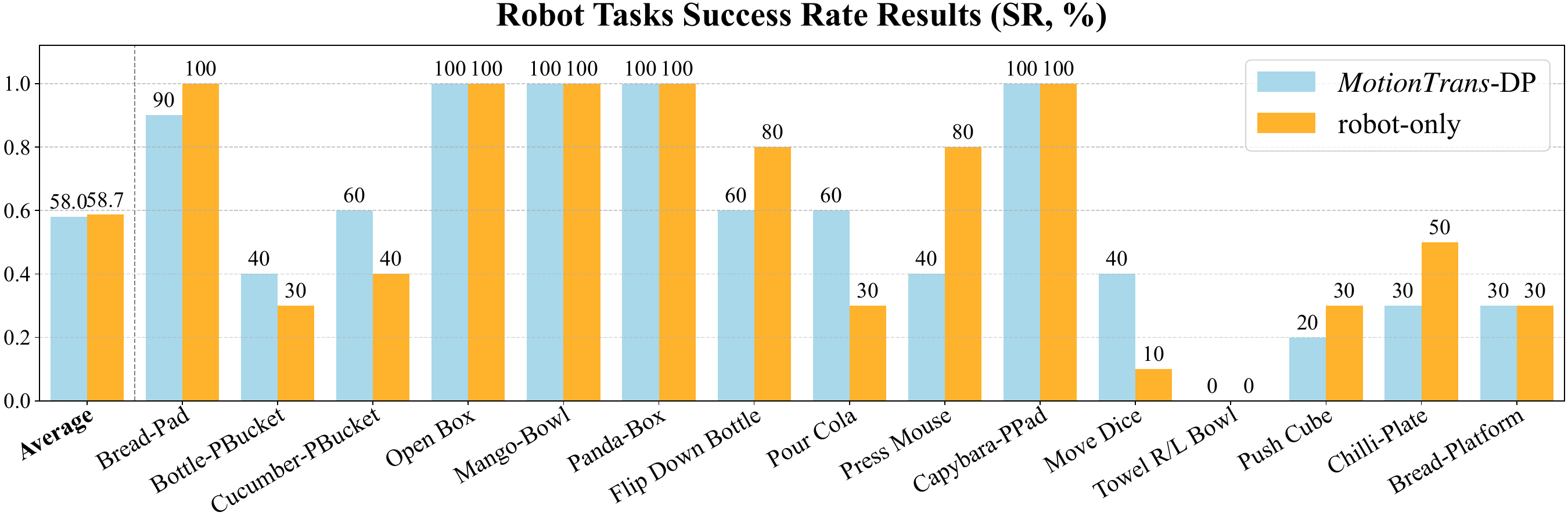}
    \caption{Results of the success rate for zero-shot experiments on robot tasks. We conclude that when considering motion learning, cotraining with non-overlapping human tasks does not ensure the improvement of policy's performance on robot tasks, with average success rates of 58.0\% for the robot-only model and 58.7\% for the \textit{MotionTrans}-DP.}
    \label{fig:zero-shot-result-robot}
\end{figure*}

In this section, we assess whether cotraining with human data can enhance task performance on robot data. To do this, we compare the Diffusion Policy (DP) trained on the complete \textit{MotionTrans Dataset} (\textit{MotionTrans}-DP) with a version trained solely on the robot data (robot-only). It is important to note that the human tasks does not overlap with robot tasks, which distinguishes our approach from previous cotraining works \cite{kareer2024egomimic, qiu2025humanoidhuman, niu2025human2locoman}. The results for all 15 robot tasks are displayed in Figure~\ref{fig:zero-shot-result-robot}. Our findings indicate that, in our setting, cotraining with non-overlapping human tasks does not significantly impact the policy's performance on robot tasks, with average success rates of 58.0\% for the robot-only model and 58.7\% for the \textit{MotionTrans}-DP. We believe this is due to the following reasons: (1) We have already collected sufficient robot demonstrations for all 15 robot tasks in our datasets, which reduces the impact of cotraining (similar to the results in Section~\ref{sec:experiment_few_shot}). (2) Although we have conducted data alignment (Section~\ref{sec:motiontrans_data_processing}), the motions from non-overlapping tasks still differ too much from those in the robot data, thereby providing insufficient auxiliary guidance for motion learning.

However, performance differences are observed in specific tasks. For example, in the ``Pour Cola" task, cotraining with human data improved the policy's grasping position, resulting in more stable pouring and better performance. Conversely, for the ``Press Mouse" task, cotraining negatively affected the final ``press" action, leading to instances where the policy only made contact without pressing. Tasks like ``Towel R/L Bowl" exhibit low success rates due to insufficient height generalization capabilities of our policies, a limitation we discuss in the limitations section  (Section~\ref{sec:conclusion}).

\subsection{Visual Background Robustness}~\label{sec:discussion_visual_robustness}
Finally, we verify the visual robustness of our experiment results against scene background~\cite{yang2025egovla}. We change the background from our default ``green table" scenes (mentioned in the dataset part in Section~\ref{sec:experiment_setup}) to a new scene, as shown in Figure~\ref{fig:main_robustness}, and evaluate Diffusion Policy (DP) performance for both zero-shot and 20-shot settings. The results are averaged across all 13 evaluation human tasks and shown on the right side of Figure~\ref{fig:main_robustness}. We observe that although the performance drops slightly, it still maintains a non-trivial Motion Progress Score and success rate. This proves the robustness of our results on motion-level human data learning. Note that this does not mean we achieve in-the-wild manipulation ability~\cite{tao2025dexwild}, which is not the main focus of this paper and will be discussed in the limitations section.

\newpage

\ 

\newpage

\begin{table}[H]
    \centering
    \renewcommand{\arraystretch}{1.25} 
    \setlength{\tabcolsep}{4pt} 
    \begin{tabular}{|l|p{5.2cm}|}
        \hline
        \textbf{Human Tasks} & \textbf{Description / VLA-prompt} \\
        \hline
        Unplug Charger & unplug the white charger. \\
        \hline
        Bread-Bucket & drop bread to the green bucket. \\
        \hline
        Press Stapler & press the stapler. \\
        \hline
        Orange-Bucket & put orange to the green bucket. \\
        \hline
        Wipe Towel & wipe blue towel on the table and push it to the bulky bottle. \\
        \hline
        Close Laptop & close silver laptop. \\
        \hline
        Mango-Bowl (Bypass) & put mango to pink bowl while avoiding obstacle by bypassing. \\
        \hline
        Mango-Bowl (Lift) & put mango to the pink bowl while avoiding obstacle by lifting. \\
        \hline
        Press Dice & press red dice to make it rotation. \\
        \hline
        Banana-Plate & put banana to the white plate. \\
        \hline
        Pour Bottle & pour bottle to the pink bowl. \\
        \hline
        Toy Bear-Box & put toy bear to the black box. \\
        \hline
        Open Box + Pand-Box & first open the white cap style box then put toy panda to the box. \\
        \hline
        Fold Towel & fold the blue towel. \\
        \hline
        Pour Milk Bottle & pour milk bottle to the yellow pan. \\
        \hline
    \end{tabular}
    \vspace{2mm}
    \caption{All 15 human tasks with detailed descriptions ($\pi_0$-VLA-prompt).}
    \label{tab:app_human_tasks}
\end{table}

\begin{table}[b]
    \centering
    \renewcommand{\arraystretch}{1.25} 
    \setlength{\tabcolsep}{4pt} 
    \begin{tabular}{|l|p{5.5cm}|}
        \hline
        \textbf{Robot Tasks} & \textbf{Description / VLA-prompt} \\
        \hline
        Push Cube & push orange cube to the bulky bottle. \\
        \hline
        Panda-Box & put toy panda to the box. \\
        \hline
        Bread-Pad & put bread to the red pad. \\
        \hline
        Open Box & open the white cap style box. \\
        \hline
        Bottle-PBucket & drop black bottle to purple bucket. \\
        \hline
        Pour Cola & pour cola to the red cup. \\
        \hline
        Move Dice & move red dice to the bulky bottle. \\
        \hline
        Flip Down Bottle & flip down the black bottle. \\
        \hline
        Press Mouse & press the pink mouse. \\
        \hline
        Bread-Platform & put bread to the high black platform. \\
        \hline
        Capybara-PPad & put Capybara to the purple pad. \\
        \hline
        Chilli-Plate & put chilli to the white plate. \\
        \hline
        Towel R/L Bowl & wipe blue towel on the table and push it left or right to the pink bowl. \\
        \hline
        Mango-Bowl & put mango to the pink bowl. \\
        \hline
        Cucumber-PBucket & put cucumber to purple bucket. \\
        \hline
    \end{tabular}
    \vspace{2mm}
    \caption{All 15 robot tasks with detailed descriptions ($\pi_0$-VLA-prompt).}
    \label{tab:app_robot_tasks}
\end{table}

\begin{table}[t]
    \centering
    \renewcommand{\arraystretch}{1.25} 
    \setlength{\tabcolsep}{4pt} 
    \begin{tabular}{|l|p{5.2cm}|} 
        \hline
        \textbf{Human Tasks} & \textbf{Rubrics of Motion Progress Score} \\ \hline
        Mango-Bowl (Bypass)  & (1) show reach-grasp; (1) successful grasp; (2) show bypassing; (2) successful bypassing; (1) show reach-put; (1) successful put;           \\ \hline
        Mango-Bowl (Lifting) & (1) show reach-grasp; (1) successful grasp; (1) show lifting; (2) successful lifting; (2) show down-putting; (1) successful put;            \\ \hline
        Pour Bottle          & (1) show reach-grasp; (1) successful grasp; (2) show rotation; (2) successful pouring; (2) good pour position;                           \\ \hline
        Toy Bear-Box        & (2) show reach-grasp; (2) successful grasp; (2) show reach-put; (1) successful put; (1) good put position;                                        \\ \hline
        Bread-Bucket        & (1) show reach-grasp; (1) successful grasp; (2) show reach-put; (2) successful put; (2) good put height;                                 \\ \hline
        Close Laptop        & (2) show reach-press; (2) press finish $<$ 30 degrees; (2) press finish $<$ 15 degrees; (2) press finish = 0 degrees;        \\ \hline
        Press Stapler       & (2) show reach-press; (2) success contact; (2) good contact position; (2) press down;                                                       \\ \hline
        Unplug Charger      & (2) show reach-grasp; (1) successful grasp; (1) show lifting; (2) successful unplug; (2) still holding after unplugging;                     \\ \hline
        Open Box + Panda-Box & (2) open the white box; (1) continue to move; (1) no stop after open the box; (1) reach the panda; (1) successful grasp the panda; (2) successful put;       \\ \hline
        Wipe Towel          & (2) show reach-press; (2) successful press; (2) show pushing (including retry); (2) successful pushing;                                      \\ \hline
        Banana-Plate        & (1) show reach-grasp; (2) successful grasp; (2) show reach-put; (2) successful put; (1) good put height;                                 \\ \hline
        Orange-Bucket       & (1) show reach-grasp; (2) successful grasp; (2) show reach-put; (2) successful put; (1) good put height;                                 \\ \hline
        Press Dice          & (1) show reach-press; (1) successful contact; (2) show press; (2) press $>$ 5 cm; (2) successful press to make it rotate; \\ \hline
    \end{tabular}
    \vspace{2mm}
    \caption{The rubrics of Motion Progress Score for all 13 evaluation human tasks. The scores are allocated to the different motions / stages required to complete the task, with a maximum score of 8 points. The number in () is the score of that stage. The ``show reach-\{action\}" rubric means policy shows approaching motion to achieve \{action\}.}
    \label{tab:motion_progress}
\end{table}

\newpage

\end{document}